\documentclass[10pt,twocolumn,letterpaper]{article}

\usepackage{cvpr}
\usepackage{times}
\usepackage{epsfig}
\usepackage{graphicx}
\usepackage{amsmath}
\usepackage{amssymb}
\usepackage{comment}
\usepackage{booktabs}
\usepackage{bbm}
\usepackage{xcolor}
\usepackage{pifont}

\definecolor{mygreen}{rgb}{0,.5,0.1}

\newcommand{\xmark}{\ding{55}}%
\usepackage[skip=10pt]{caption}

\newcommand{\nofootnote}[1]{}

\usepackage[pagebackref=true,breaklinks=true,colorlinks,bookmarks=false]{hyperref}

\cvprfinalcopy 

\def\cvprPaperID{1133} 

\ifcvprfinal\fi

\title{TransFill: Reference-guided Image Inpainting by\\ Merging  Multiple Color and Spatial Transformations}

\author{
Yuqian Zhou$^1$, Connelly Barnes$^{2}$, Eli Shechtman$^2$, Sohrab Amirghodsi$^2$ \\
$^1$IFP, UIUC,~~ $^2$Adobe Research
}

\begin{document}

\twocolumn[{%
\renewcommand\twocolumn[1][]{#1}%
\maketitle
\vspace{-6ex}
\begin{center}
    \centering
    \includegraphics[width=\textwidth]{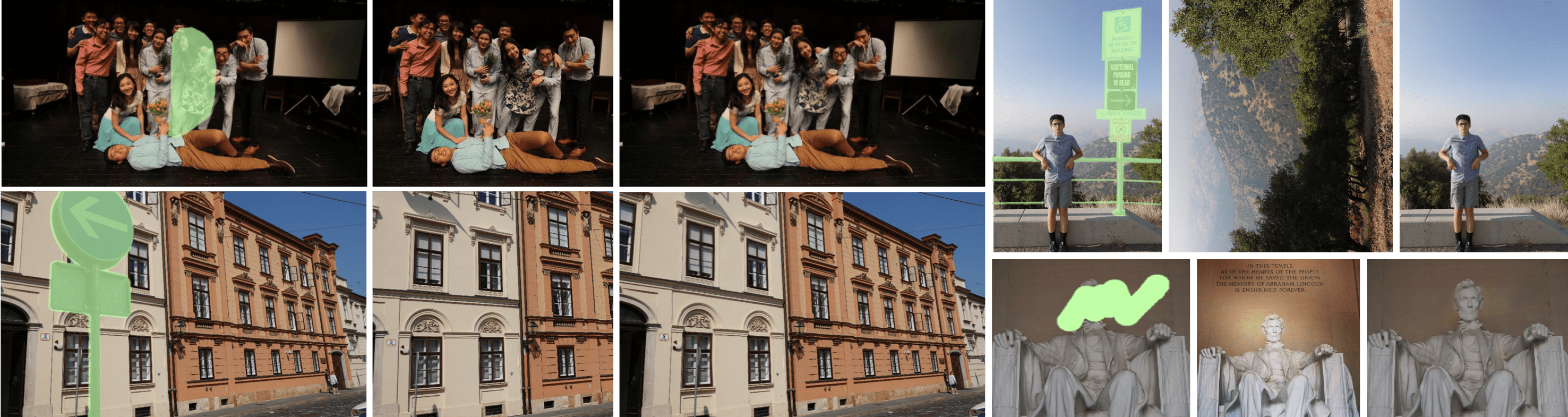}
    \captionof{figure}{Results of our reference-guided inpainting for user-provided images. We show multiple practical applications like replacing and removing foreground people and objects. Each triad shows the target image with the hole, the source image used as a reference, and the inpainting result. Our method has strong performance and addresses challenging real-world issues such as parallax, 90 degree image rotations, and lighting inconsistency between the source and target images.} 
\end{center}%
}]

\begin{abstract}
Image inpainting is the task of plausibly restoring missing pixels within a hole region that is to be removed from a target image. Most existing technologies exploit patch similarities within the image, or leverage large-scale training data to fill the hole using learned semantic and texture information. However, due to the ill-posed nature of the inpainting task, such methods struggle to complete larger holes containing complicated scenes. In this paper, we propose TransFill, a multi-homography transformed fusion method to fill the hole by referring to another source image that shares scene contents with the target image. We first align the source image to the target image by estimating multiple homographies guided by different depth levels. We then learn to adjust the color and apply a pixel-level warping to each homography-warped source image to make it more consistent with the target. Finally, a pixel-level fusion module is learned to selectively merge the different proposals. Our method achieves state-of-the-art performance on pairs of images across a variety of wide baselines and color differences, and generalizes to user-provided image pairs. More details are available at the \href{https://yzhouas.github.io/projects/TransFill/index.html}{project website}.
\end{abstract}


\section{Introduction}
Image inpainting is an image restoration task where the goal is to fill in specific regions of the image while making the entire image visually realistic. The regions to be filled are called hole regions, and could contain undesired foreground objects or small distracting elements, or corrupted regions of the image. Much research has been devoted to improving image inpainting either by image self-similarity (e.g. \cite{barnes2009patchmatch}) or deep generative models (e.g.~\cite{yu2019free,xiong2019foreground,yu2018generative}). Such methods synthesize realistic semantics and textures by reusing similar patches from non-hole regions or learning from large collections of images, respectively. However, those methods still struggle in cases when holes are large, or the expected contents inside hole regions have complicated semantic layout, texture, or depth. 

These problems can be addressed if there happens to be a second reference image of the same scene that exposes some desired image content that can be copied to the hole. This task is referred to as \emph{reference-guided} image inpainting in the literature~\cite{oh2019onion}, but this topic is less explored. In our paper, we call the image with the hole indicated for removal the \emph{target image}. In general, there could be multiple other \emph{source images} used as references. These could be taken by the photographer for the same scene after objects have moved or the photographer moved the camera to a different viewpoint to expose the background. Alternatively, a source image could be collected from the Internet \cite{WSZ09}. If one such source image contains new desired appearance for the target hole region, then we can copy the pixels from the source to fill in the target hole regions. In this paper we assume that the user has identified a particular source image with the new desired appearance, so we refer to this as \emph{the source} image. We imagine that dedicated apps might be created for aiding the photographer in this process, or for automatically retrieving suitable such source images from the Internet. 

Although the reference source image makes the inpainting task easier, reference-guided inpainting is still quite challenging for several reasons. First, the hole regions could be very large, which makes the task of guessing the pixel colors in the hole region less well-posed. Second, we wish for our task to be as general as possible, so we allow an uncalibrated camera to freely translate to different 3D positions for the source and target image, because this can allow the photographer to intentionally reveal regions behind a foreground object to be removed. Such translations, however, can induce large parallax, which cannot be modeled in image space by a simple 2D warp such as a global homography. Unlike video inpainting or multi-view Structure-from-Motion (SfM), we assume the system will not have access to more than two photos. Thus, it is harder in our setting to reliably estimate 3D structures, depth, and point correspondences. Third, depending on the camera and photography setup, the photographs may have substantially different exposure, white balance, or lighting environment, and if one photograph comes from the Internet, then it will have different camera response curves. Existing methods based purely on warping cannot resolve the resulting complex issues of color matching. Finally, there may exist  regions in the source image that do not exist after warping due to pixels being out of the image or occluded. 

To address these challenges, we propose a multi-homography fusion pipeline combined with deep warping, color harmonization, and single image inpainting. We address the issue of parallax by assuming that there may be multiple depth planes inside the hole. Loosely inspired by recent work on multiplane images~\cite{flynn2016deepstereo,zhou2018stereo,mildenhall2019local,tucker2020single}, we propose multiple homographic registrations of the source image to the target, each corresponding to an assumption that the scene geometry lies on a different 3D plane. Given a target and a source image, we first estimate the matched feature points between the two images, cluster the inliers according to their estimated depths in the target image, and for each cluster estimate a single homography to perform an initial image registration. We call each of these candidate alignment images a \emph{proposal}. For each proposal, we then tackle the challenge of color matching by using a deep bilateral color transformation, and we address parallax issues by refining the warp using a learned per-pixel spatial transformation. We then merge all the transformed source image proposals by learning a set of fusion masks. Finally, we address the last challenge regarding regions which do not exist in the source image by using a state-of-the-art single image inpainting method to complete missing regions, and learn to merge it as well. 

In summary, the main contributions of our method are: (1) We propose TransFill, a multi-homography estimation pipeline to obtain multiple transformations of the source image, where each aligns a specific region to the target image; (2) We propose to learn a color and spatial transformer to simultaneously perform a color matching and make a per-pixel spatial transformation to address any residual differences after the initial alignment; (3) We learn weights suitable for combining all final proposals with a single image inpainting result.

\section{Related Work}
\textbf{Image inpainting}. Inpainting research can be divided into two categories: traditional methods that work by propagating  colors or matching patches, and deep methods that learn semantics and texture from large image datasets.

Some traditional methods propagate pixel colors by anisotropic diffusion~\cite{bertalmio2000image} or solving PDEs~\cite{ballester2001filling}. Such methods work well for thin hole regions but as the hole regions grow larger they tend to result in over-blurring. Patch-based image inpainting methods work by finding  similar matches elsewhere in the image and copying the resulting texture \cite{wexler2007space, barnes2009patchmatch}. Those methods tend to result in high-quality texture but may give implausible structure and semantics.

Our work is more closely related to deep models for inpainting that use a single image. Context encoders analyze the surroundings of the hole~\cite{pathak2016context}, local and global discriminators~\cite{iizuka2017globally} can improve local texture and overall image layout, and partial~\cite{liu2018image} and gated convolutions~\cite{yu2019free} can reduce artifacts from filter responses at the hole boundary. 

More recently, some deep methods have focused on inferring other information first: these can be roughly categorized into using edges~\cite{nazeri2019edgeconnect}, segmentation masks~\cite{song2018spg}, low-frequency structures~\cite{ren2019structureflow,liao2020guidance}, and other possible maps like depth. The ill-posed nature \cite{zheng2019pluralistic} of single-image inpainting makes it challenging to complete larger holes and higher-resolution images. Recent works demonstrate neural networks can generate high-resolution images~\cite{yang2017high,zeng2020high,yi2020contextual}, but for large holes, these methods can still generate results that appear semantically implausible or have artifacts in the fine-scale texture. Since our method has a source reference image, we can better establish consistency with the ground truth image by learning appropriate spatial and color transformations for a source image patch.

\textbf{Video inpainting.} A few classical works in this area are Wexler \etal~\cite{wexler2007space} and Granados \etal~\cite{granados2012background}, which globally optimize patch-based energies, and  Newson \etal~\cite{newson2014video}, which estimates multiple homographies using a piecewise planar assumption for the scene. Xu \etal \cite{xu2019deep} estimates the optical flow to learn the pixel warping field. Recently, the Onion-Peel Network (OPN) \cite{oh2019onion} leverages non-local designs inside the network, making it feasible to apply multi-source inpainting for a larger temporal range. Lee \etal \cite{lee2019copy} proposed a Copy-and-Paste Network to learn the alignment of consecutive frames for video inpainting. Zhao \etal \cite{zhao2019guided} reuse contents from an unrelated image for a reference-based inpainting. Their method is based on only a single affine transform, which we show is not enough in our experiments, and exhibits residual color and geometric incompatibilities that are problematic in our multi-view scenario. Xue \etal \cite{xue2015computational} is a specialized method designed to remove reflective or occluding elements near the camera such as fences.

\textbf{Image harmonization.}
Image harmonization refers to matching the color distribution and appearance when compositing a foreground from one image on a background from another image. Traditional methods transfer color statistics locally and globally \cite{pitie2005n, reinhard2001color,sunkavalli2010multi} and use gradient-domain based blending \cite{perez2003poisson,jia2006drag,tao2010error}. Digital photomontage~\cite{agarwala2004interactive} also demonstrated copy-and-paste workflows that can change the appearance of a foreground subject. Unlike our method, photomontage required user input and assumes the photographs have been aligned. 
Recently, CNN-based harmonization models \cite{zhu2015learning, xiaodong2019improving} are emerging, including methods involving segmentation masks \cite{tsai2017deep} for region selection, and discriminators for domain verification \cite{cong2020dovenet}. Deep bilateral filtering has also been used to better preserve edges and details while transforming image color space~ \cite{gharbi2017deep,wang2019underexposed}. Our work is the first to integrate harmonization with a neural network for reference-guided inpainting. We apply a deep bilateral color transformation to address color inconsistencies while preserving edges. 

\textbf{Image alignment.}
Image alignment or registration involves placing multiple images in the same coordinate system. It is widely used for video stabilization \cite{liu2016meshflow}, image stitching \cite{lee2020warping,herrmann2018robust}, and serves as an important pre-processing step for many video and image applications like face analysis. Homography warping is a widely used global parametric method. Sparse local features like SIFT~\cite{lowe1999object} can be matched either using nearest neighbour, or deep models like OANet \cite{zhang2019oanet} and SuperGlue \cite{sarlin2020superglue}, and the resulting correspondences can be used to estimate warping models. Recently, deep models have been explored to directly learn  homography parameters \cite{detone2016deep,zhang2019content,nguyen2018unsupervised}, demonstrating their advantages on low-light and low-texture images. 

Issues of parallax due to content at different depths can be better addressed by mesh-based warping \cite{liu2016meshflow,li2015dual,liu2009content} or pixel-wise dense optical flow \cite{horn1993determining,wulff2015efficient,sun2010secrets,sun2018pwc,weinzaepfel2013deepflow,ilg2017flownet}. Liu \etal proposed the Content Preserving Warp (CPW) \cite{liu2009content} to maintain the rigidity of motions. Recently, Ye \etal proposed deep meshflow \cite{ye2019deepmeshflow} to make  mesh estimation more robust on different scenes. Due to the sparsity of the mesh, image contents can be better retained while warping. However, optical-flow based methods can provide greater flexibility in permitted motions.  
Our pipeline uses multiple global homographies followed by per-pixel warping fields to combine the advantages of various alignment methods.

\vspace{-1.5ex}
\section{Method}
\begin{figure}[t]\setlength{\belowcaptionskip}{-10pt}
\centering
  \includegraphics[width=\linewidth]{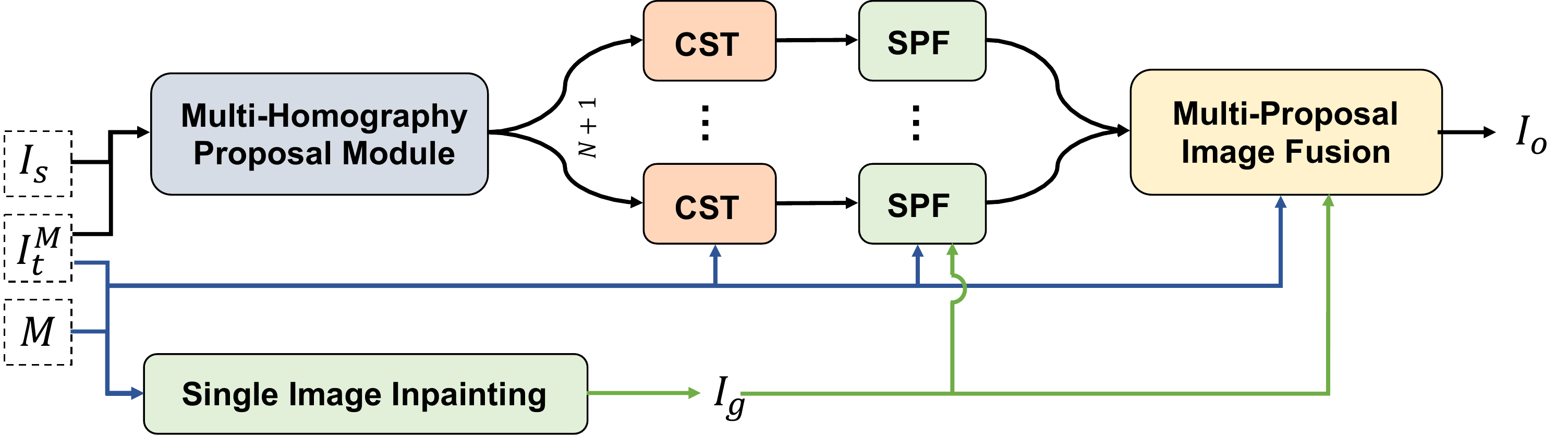}
\caption{System pipeline. Given the target image $I_t^M$ masked by an associated binary hole image $M$, and a single source image $I_s$, we first propose multiple global homographies using the multi-homography  proposal module, and locally adjust color and spatial misalignments in each proposal using our Color-Spatial Transformer (CST). Then we merge each proposal with the output $I_g$ from a single-image inpainting model using Single-Proposal Fusion (SPF), and finally selectively blend all the proposals. }
  \label{fig:sys}
\end{figure}
We will first give an overview of our pipeline. Suppose we are given a target image $I_t \in {\rm I\!R}^{W \times H \times 3}$, an associated mask $M \in {\rm I\!R}^{W \times H \times 1}$, and a single source image $I_s \in {\rm I\!R}^{W_s \times H_s \times 3}$. Note that $M$ indicates the hole regions with value one, and elsewhere with zero. The masked target image is then denoted by $I_t^M = (1-M) \odot I_t$. We assume there is sufficient overlap in content between the two images especially nearby (but not necessarily within) the masked regions. Our task is to generate contents inside the masked regions of $I_t$ by effectively reusing contents of $I_s$. More specifically, we wish to geometrically align  $I_s$ with $I_t$ in the vicinity of the hole region globally and locally, and adjust any color inconsistency. We fill any regions that are occluded or outside the image using a state-of-the-art single image inpainting method. 

Our pipeline follows four steps as shown in Figure \ref{fig:sys}. It includes an initial registration using multiple homography proposals, per-pixel color and spatial transformations for each proposal, single-proposal fusion and multi-proposal fusion, as introduced in the following sections. 

\subsection{Multi-homography Proposals}
\begin{figure}[t]
\centering
  \includegraphics[width=\linewidth]{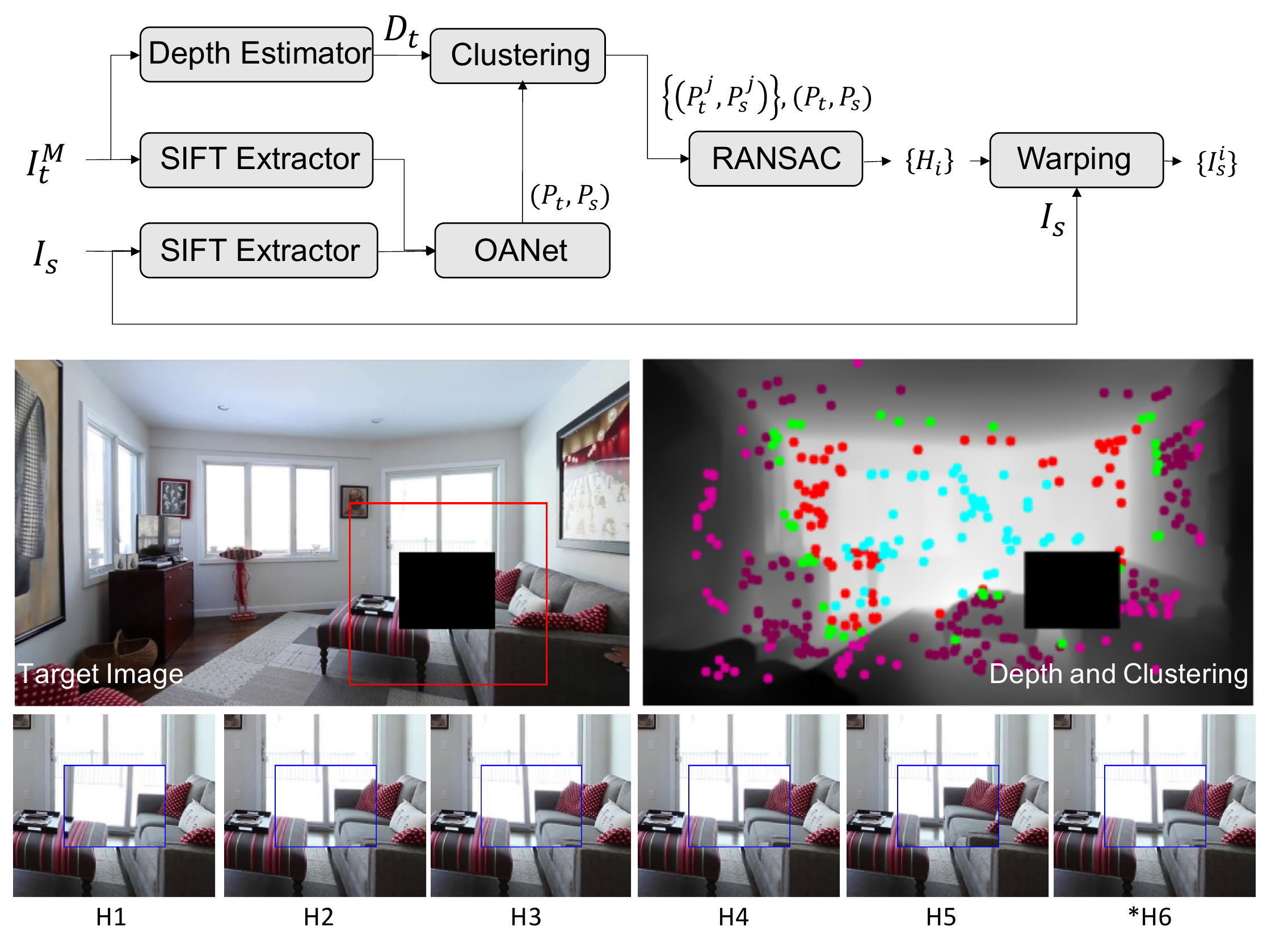}
\caption{Multi-homography Proposal Module. We compute the monocular depth $D_t$ of the non-hole region $I_t^M$, and cluster the feature matching points into $N$ sub-groups using the depth values. Each estimated homography $H_i$ will align different regions within the hole. $^*H_6$ indicates a homography estimated using all the points.}
  \label{fig:mh}
  \vspace{-6mm}
\end{figure}
In this stage, we first globally warp the source image $I_s$ to align it with the masked target image $I_t^M$. Provided the contents inside the hole region occur at multiple depth planes, or the camera motion is not a simple rotation, a single homography is not sufficient to perfectly align the source and target image \cite{gao2011constructing}. Therefore, we propose to estimate multiple homography matrices to transform $I_s$. Ideally, each homography-transformed $I_s$ can align with $I_t$ within a specific image depth level range or local spatial area, as shown in Figure \ref{fig:mh}.

To obtain different transformation matrices, we first extract SIFT~\cite{lowe2004distinctive} features from $I_t^M$ and $I_s$, and feed all the extracted feature points and their descriptors into a pre-trained OANet~\cite{zhang2019oanet} for outlier rejection. The lightweight OANet efficiently establishes the correspondences between $I_t^M$ and $I_s$ by considering the order of the points in the global and local context. OANet outputs the inliers forming a point set $P_t$ in $I_t^M$, and its corresponding matched point set $P_s$ in $I_s$. 
To consider different possible depth planes within and nearby the hole region, we are inspired by the Multi-Plane Image (MPI) \cite{zhou2018stereo} idea for scene synthesis. We estimate the depth map $D_t$ from $I_t^M$ using a deep learning based monocular depth estimator~\cite{Hu2019RevisitingSI} , and record the depth value for each point in $P_t$. We then cluster those points into a partition of $N$ subsets $\{P_t^j\}, j\in{[1, N]}$ by their depth values using an agglomerative clustering method \cite{johnson1967hierarchical}, where $P_t = \cup^{N}_{j=1} P_t^j$. The corresponding matched points in $P_s$ are used to form the subsets $P_s = \cup^{N}_{j=1} P_s^j$ accordingly.

For each subset's pairs of points $(P_t^j, P_s^j)$, we estimate a single homography using RANSAC \cite{fischler1981random}. By further including the homography estimated from the full set of points $(P_t, P_s)$, we obtain $N+1$ homography matrices overall. We denote them by $H_i, i\in{[1,N+1]}$. 
Finally, we transform the source image $I_s$ using the estimated $H_i$, and obtain a set of warped source images $\{I_s^i\}$, where $I_s^i \in {\rm I\!R}^{W \times H \times 3}, i\in{[1,N+1]}$. We set $N=5$ in our experiments.

\subsection{Color-Spatial Transformation (CST) Module}
\begin{figure}[t]\setlength{\belowcaptionskip}{-10pt}
\centering
  \includegraphics[width=\linewidth]{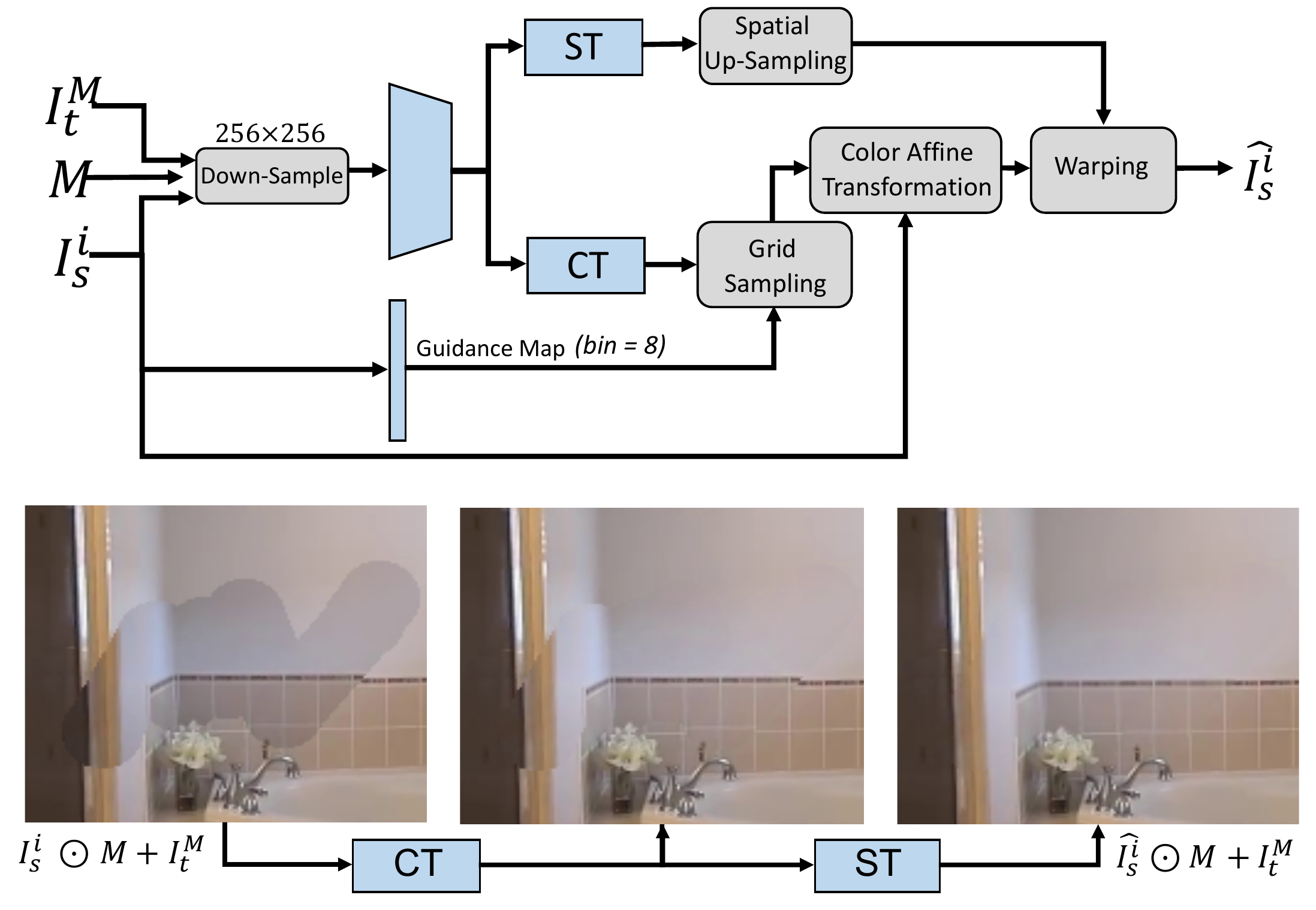}
\caption{Structure of the Color-Spatial Transformer Module. $I_s^i$ will first go through a Color Transformer (CT), and then a Spatial Transformer (ST) to obtain a refined source image $\hat{I_s^i}$. The bottom row shows examples of the refinement stages. Blocks with blue color indicate there are learned parameters, otherwise they are parameter-free.}
  \label{fig:cst}
\end{figure}

The global homography-warped source image sets $\{I_s^i\}$ are regarded as the initialization of the warping of $I_s$. However, as shown in Figure \ref{fig:mh} and \ref{fig:cst}, while directly compositing $I_s^i$ and $I_t^M$ using $ I_s^i \odot M + I_t^M$, due to the possibly inaccurate homography estimation or challenges of large parallax, there may be small misalignments inside and near the hole region, especially along the hole boundary. Additionally, the composite image may suffer from color and exposure differences. Therefore, we propose another refinement step that we call a Color-Spatial Transformer (CST). This simultaneously adjusts the color and alignment for each $I_s^i$. The structure of CST is illustrated in Figure \ref{fig:cst}. $I_s^i$ will first go through a Color Transformer (CT), and then a Spatial Transformer (ST) to obtain a refined source image $\hat{I_s^i}$.

In our design of the color and spatial transformers, we would like to retain the texture details and the rigidity of the source image contents. Additionally, we prefer the color transformation and warping operations to be decoupled and not have to use auxiliary losses for each component. Inspired by deep bilateral filtering \cite{gharbi2017deep} and Spatial-Transformer Network (STN) \cite{jaderberg2015spatial}, we propose to learn the transformations in a lower resolution, and obtain the full-resolution coefficients using up-sampling. Specifically, given $I_s^i$, $I_t^M$ and $M$, we down-sample them to $256 \times 256$ to obtain $I_s^i \downarrow$, $I_t^M\downarrow$ and $M\downarrow$. Then we compute the high-level features $u_s^i = B(I_s^i \downarrow, I_t^M \downarrow, M \downarrow)$ using a shared network $B$. After that, the color and spatial transformation coefficients will be learned by the CT and ST sub-networks. 

\textbf{Color Transformation (CT).}
To transform the color in RGB space of $I_s^i$ to $I_{sc}^i$, we learn an affine transformation with parameters $A_c^i = [K_c^i \quad b_c^i] \in {\rm I\!R}^{W\times H \times 3 \times 4}$. Formally, for each pixel at location $p$,  $I_{sc}^i(p) = K_c^i(p) I_s^i(p) + b_c^i(p)$, where $K_c^i(p) \in {\rm I\!R}^{3\times 3}$ and $b_c(p) \in {\rm I\!R}^{1\times 3}$. To better preserve the edges and textual details, we adopt deep bilateral filtering \cite{gharbi2017deep}. Specifically, we learn a bilateral grid $\bar{A_c^i} = B_c(u_s^i) \in \rm I\!R^{s \times s \times d \times 3 \times 4}$ in a lower resolution, and a single-channel guidance map $g_c^i = G_c(I_s^i) \in {\rm I\!R}^{W \times H \times 1}$ in full-resolution. We fix $s = 8$ and $d = 8$ in our experiments. $B_c$ and $G_c$ are the trainable networks for estimating the grid and guidance map. Finally, $A_c^i$ is tri-linearly sampled from $\bar{A_c^i}$ using the normalized triplet $(x, y, g_c^i(p))$.

\textbf{Spatial Transformation (ST).}
We learn the spatial warping offset $A_s^i = [A_{sx}^i\quad A_{sy}^i] \in {\rm I\!R}^{W\times H \times 2}$ along the horizontal and vertical axes. To better preserve the rigidity of the image contents inside hole region, we propose to learn the warping field $\bar{A_s^i} = B_s(u_s^i) \in \rm I\!R^{s \times s \times 2}$ in a lower resolution, and up-sample it to $A_s^i$ using bi-linear interpolation. Finally, $\hat{I_s^i} = \mathrm{Warp}(I_{sc}^i; A_s^i)$. The objective loss to learn the CST module is defined by,
\vspace{-1ex}
\begin{equation}
    \mathcal{L}_{CS}^i = || M_v \odot M \odot (I_t - \hat{I_s^i})||_1,
\end{equation}

where $M_v = \mathbbm{1}(I_s^i >0)$ is the valid mask indicating the pixel regions after initial homography warping. 

\subsection{Single-Proposal Fusion (SPF) Module}
\begin{figure}[t]\setlength{\belowcaptionskip}{-10pt}
\centering
  \includegraphics[width=0.85\linewidth]{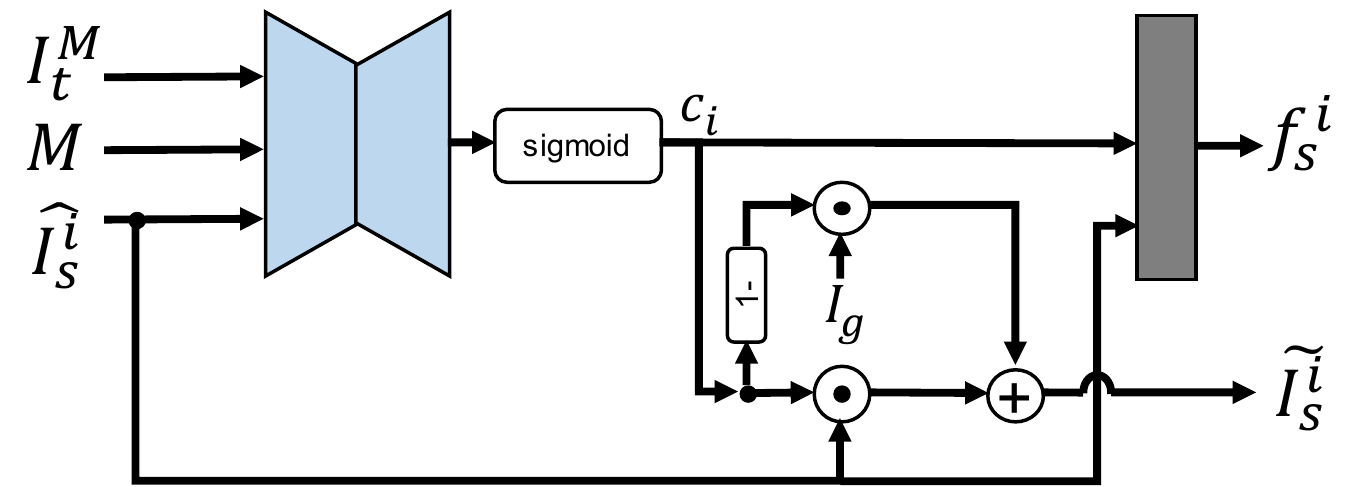}
\caption{Single-Proposal Fusion (SPF) module. This takes $I_t^M$, $M$, a single $\hat{I_s^i}$ and $I_g$ as inputs, where $I_g$ is the result of a single image inpainting method. SPF outputs a confidence map $c_i$, the merged $\tilde{I_s^i}$, and the packed features $f_s^i$.}
  \label{fig:spf}
\end{figure}

The Single-Proposal Fusion (SPF) module learns to estimate a confidence map and other features for the refined results $\hat{I_s^i}$ from the CST module by merging it with the outputs of a well-trained single image inpainting model called ProFill~\cite{zeng2020high}. The inpainting results from ProFill often generate good structures, so the intuition for the SPF module is that we independently do an image comparison of each proposal against this ProFill reference, to better constrain the overall learning task and learn confidence and difference features that can help the harder downstream multi-proposal fusion task. As shown in Figure \ref{fig:spf}, the module takes $I_t^M$, $M$, a single $\hat{I_s^i}$ and $I_g$ as inputs, where $I_g$ is the output from a single image inpainting method. In this paper, we use a pre-trained ProFill~\cite{zeng2020high} model and freeze its weights while training the whole pipeline. The module outputs a confidence map of $\hat{I_s^i}$ denoted by $c_i$ of the same spatial size as $I_t^M$. The output $\tilde{I_s^i}$ of merging is
\vspace{-1ex}
\begin{equation}
    \tilde{I_s^i} = c_i \odot \hat{I_s^i} + (1 - c_i) \odot I_g, 
\end{equation}

The values in the confidence map range from zero to one, and higher-valued regions should contain more informative and realistic pixels. The composited result of merging $I_t^M + M \odot \tilde{I_s^i}$ can also be displayed to the user as an intermediate result demonstrating the performance of a single proposal. In our experiments, we will show that compared to learning to merge multiple $\hat{I_s^i}$ directly, it is better to condition on the outputs from a well-learned SPF module.

Additionally, we utilize a shallow convolutional module to concatenate the learned confidence map $c_i$ and the output of the CST module $\hat{I_s^i}$, and output a three-channel feature map $f_s^i$ to be fed into the final multi-proposal fusion module in section \ref{sec:mpf}. Similarly, when we input $I_g$ to the SPF, we obtain the feature $f_g$. The objective function for learning the SPF is defined as,
\begin{equation}
    \mathcal{L}_E^i = || M \odot (I_t - \tilde{I_s^i})||_1,
\end{equation}
and an additional Total Variance loss is imposed on $c_i$ to enforce the smoothness of the map.
\begin{equation}
    \mathcal{L}_c^i = \mathcal{L}_{\mathrm{TV}}(c_i), \mathcal{L}_{\mathrm{TV}}(u) = \left\lVert \frac{\partial u}{\partial x}  \right\rVert_1 + \left\lVert \frac{\partial u} {\partial y} \right\rVert_1
\end{equation}
\vspace{-2ex}
\subsection{Multi-Proposal Fusion (MPF) Module}\label{sec:mpf}
\begin{figure}[t]\setlength{\belowcaptionskip}{-10pt}
\centering
  \includegraphics[width=0.85\linewidth]{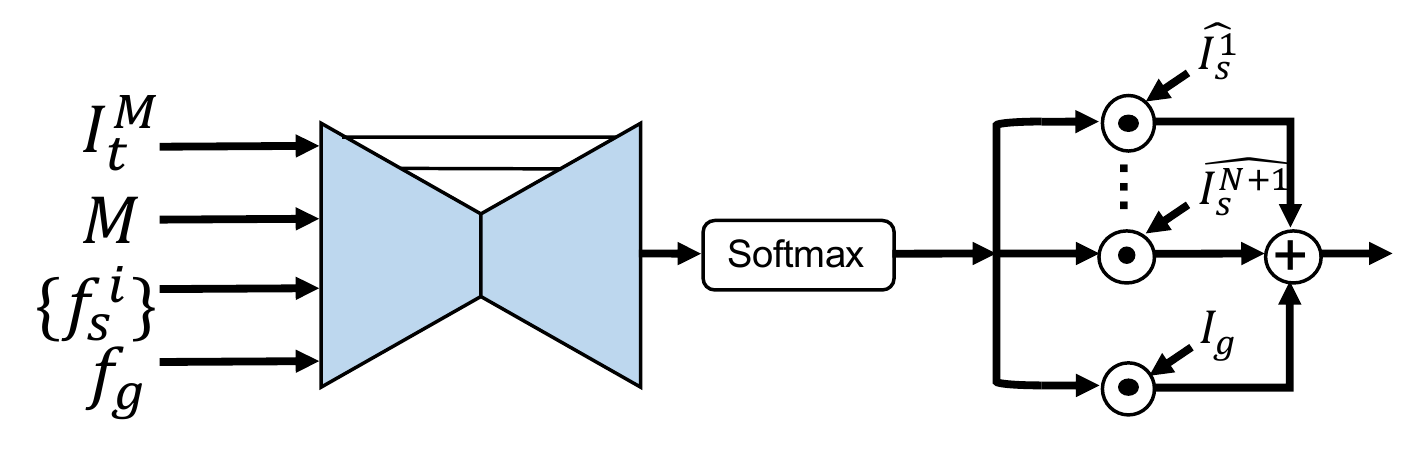}
\caption{Structure of the Multi-Proposal Fusion (MPF) Module. We feed the UNet with packed features $f_s^i$ and $f_g$ from the SPF module, and learn a spatially-varying  merging mask for all the proposals.}
  \label{fig:mpf}
\end{figure}
The Multi-Proposal Fusion (MPF) module merges the $N+1$ proposals of the refined source images $\hat{I_s^i}$ and the single-image inpainting results $I_g$ together. The module is fed with the packed features $f_s^i$ and $f_g$ from the SPF module. Pixel-wise merging weights ${\overline{c_i}}, i \in [1,N+1]$ and $c_g$ are learned through a UNet \cite{ronneberger2015u} with softmax ($c_g + \sum_{i=1}^{N+1}\overline{c_i} = 1$) by merging different portions of proposals as $I_m$,
\begin{equation}
    I_m = c_{g}\odot I_g + \sum_{i=1}^{N+1} {\overline{c_i} \odot \hat{I_s^i}},
\end{equation}

Then the final result $I_o = I_t^M + M \odot I_m$ is learned by the objective functions,
\begin{equation}
    \mathcal{L}_o = || M \odot (I_t - I_o)||_1 + VGG( M \odot I_t,  M \odot I_o),
\end{equation}

\noindent where the VGG loss matches features of the pool5 layer of a pre-trained VGG19 \cite{simonyan2014very}. Similarly, total variance losses are imposed to the weighting maps $\overline{c_i}$ and $c_g$, so we have the losses $\mathcal{L}_{\overline{c}}^i = \mathcal{L}_{\mathrm{TV}}(\overline{c})$ and $\mathcal{L}_c^g = \mathcal{L}_{\mathrm{TV}}(c_g)$. Therefore, the overall loss function with $\lambda_1 = 1, \lambda_2 = 1$ becomes
\vspace{-1ex}
\begin{equation}
\resizebox{.9\hsize}{!}{
    $\mathcal{L}_{all} = \mathcal{L}_{o} + \lambda_1  \mathcal{L}_{c}^g + \sum_{i=1}^{N+1} {(\mathcal{L}_{CS}^i + \mathcal{L}_{E}^i + \lambda_2 (\mathcal{L}_{c}^i + \mathcal{L}_{\overline{c}}^i) )}.
    $ }
\end{equation}

\section{Experimental Results}
\subsection{Datasets and Implementation}
\textbf{Datasets.}
We trained the model on the RealEstate10K dataset~\cite{zhou2018stereo}. This was collected from YouTube videos labelled as real estate footage. In total it consists of more than 8000 video clips with length from 1 to 10 seconds. For each clip, we randomly sampled pairs of images with a displacement of 10, 20, and 30 frames apart. We call this ``Frame Displacement" (FD). This resulted in 188184 frame pairs for training, and 20290 pairs for testing. We generated random free-form brush-and-stroke holes like in DeepFillv2 \cite{yu2019free}. We also collected 3K more pairs of real user-provided image pairs to serve as practical user cases for testing. 

For training the Color-Spatial Transformer (CST), although RealEstate10K contains sufficient samples with real multi-view data and different exposures across image pairs, it lacks image pairs with large color inconsistency. Therefore, we synthesized misaligned color-different images from the MIT-Adobe5K dataset \cite{fivek}, and uniformly mixed these data with RealEstate10K for training. Adobe5K contains 5000 images, and for each image it provides five additional expert-retouched images to form 5000 sets in total. We regard the original samples as target images and synthesized the misaligned source images using the method in \cite{detone2016deep}. We make two binary variables  for whether there is a color difference ($C$) and whether there is spatial misalignment ($S$), and synthesized pairs with $CS$, $C\bar{S}$, $\bar{C}S$ and $\bar{C}\bar{S}$ with equal probability from 4000 sets to form a balanced training set, leaving 1000 sets for validation. 

\textbf{Implementations.}
 We obtained a pre-trained OANet \nofootnote{OANet: https://github.com/zjhthu/OANet} model for image feature matching and outlier rejection. We applied the pretrained model\nofootnote{Hu et al.: https://github.com/JunjH/Revisiting$\_$Single$\_$Depth$\_$Estimation} of Hu \etal \cite{Hu2019RevisitingSI} to estimate the depth map from a single target image. We also obtained a pre-trained ProFill~\cite{zeng2020high} from the authors. All the above-mentioned model weights were frozen during training. Additionally, we pre-trained the CST module using the mixed dataset in advance for 400 epochs, and froze its weights afterwards. Finally, the whole pipeline was trained end-to-end for 400 more epochs. 
 We used a patch size of $256\times 256$ for training and arbitrary size for inference, and a learning rate of $10^{-4}$ with decay rate 0.5 after 200 epochs. We used the Adam optimizer~\cite{kingma2014adam} with betas (0.9, 0.999). The code is implemented in PyTorch \cite{paszke2019pytorch}.

\subsection{Baseline Models}
\begin{figure*}[t]\setlength{\belowcaptionskip}{-15pt}
\setlength{\abovecaptionskip}{0pt}
\centering
  \includegraphics[width=\linewidth]{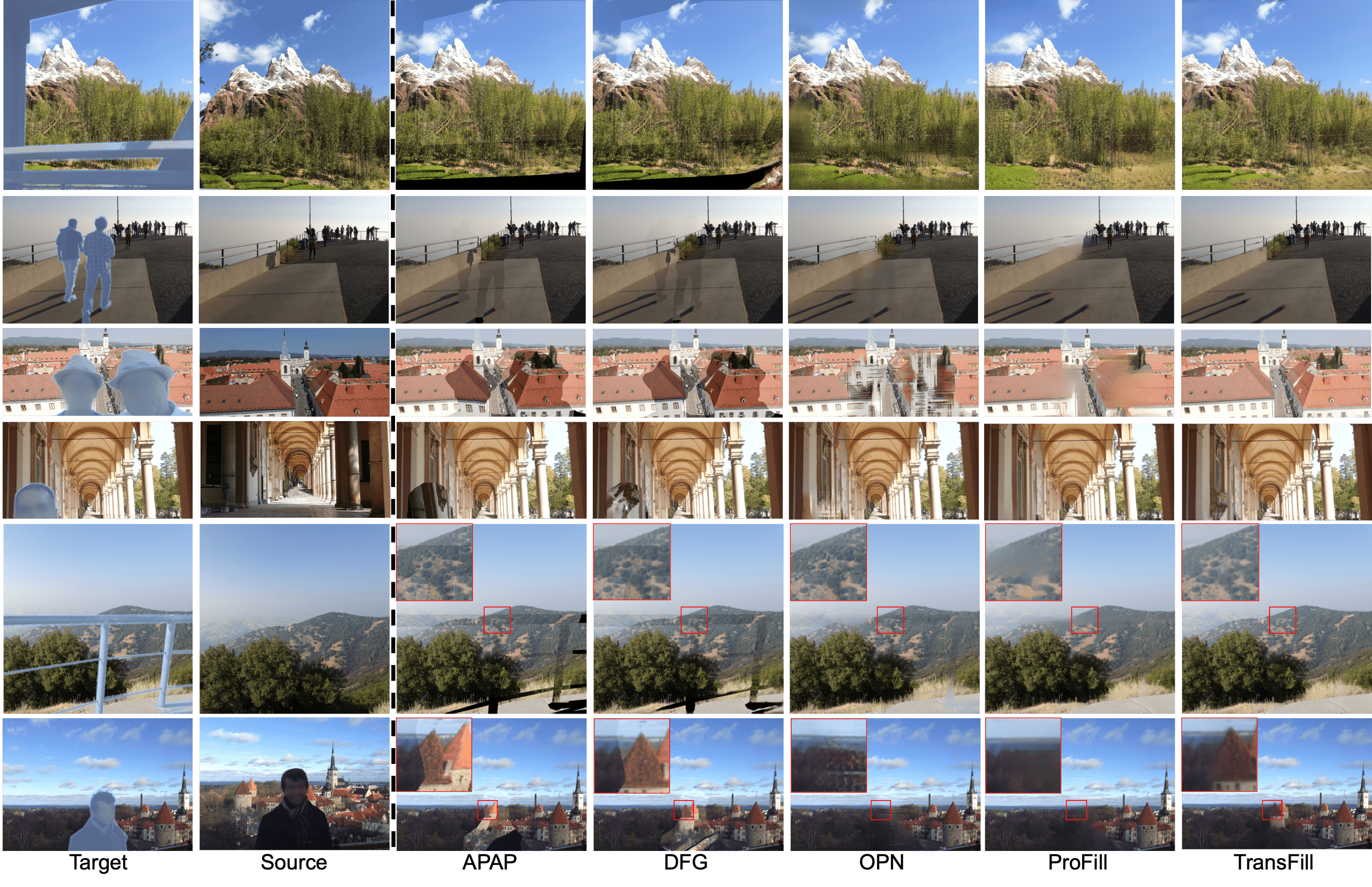}\vspace{-0.75ex}
\caption{Comparison with baselines on challenging user-provided image pairs. For better visualization, we only crop the regions of interest from the whole target and source images. Please zoom in to see the details.}\vspace{1ex} 
  \label{fig:res1}
\end{figure*}
We chose baselines that are similar to, but may not exactly the same as our task, including approaches addressing image stitching \cite{zaragoza2013projective}, optical flow-guided video inpainting \cite{xu2019deep}, non-local patch matching for multiple photo inpainting \cite{oh2019onion}, and a state-of-the-art single image inpainting method \cite{zeng2020high} with the reference image concatenated so the method has access to the same inputs as the rest. \\
\textbf{APAP} \cite{zaragoza2013projective}: As-Projective-As-Possible is a baseline image stitching algorithm that resolves depth parallax. We used the official Matlab \footnote{APAP: https://cs.adelaide.edu.au/$\sim$tjchin/apap/} implementation for testing. \\
\textbf{DFG} \cite{xu2019deep}: Deep Flow-Guided Video Inpainting treats video inpainting as pixel propagation. It fills the holes by completing the optical flow field estimated by FlowNet2.0 \cite{ilg2017flownet}. We used their official\footnote{DFG: https://github.com/nbei/Deep-Flow-Guided-Video-Inpainting} pre-trained model for testing.\\
\textbf{OPN} \cite{oh2019onion}: Onion-Peel Network is a recent work addressing video and group photo inpainting  using non-local attention blocks. We used their official PyTorch code\footnote{OPN: https://github.com/seoungwugoh/opn-demo}. \\
\textbf{ProFill} \cite{zeng2020high}: ProFill is a state-of-the-art single-image inpainting method that also contains a contextual attention module \cite{yu2018generative}. 
We used the official pre-trained model \footnote{ProFill: https://zengxianyu.github.io/iic/} from the authors. When testing, we fed in the target with the homography-warped source image. Before testing on RealEstate10K, we also fine-tuned OPN and ProFill on RealEstate10K training frames for fairness.

\subsection{Qualitative Comparison}
\begin{figure*}[t]\setlength{\belowcaptionskip}{-10pt}
\setlength{\abovecaptionskip}{0pt}
\centering
  \includegraphics[width=1\linewidth]{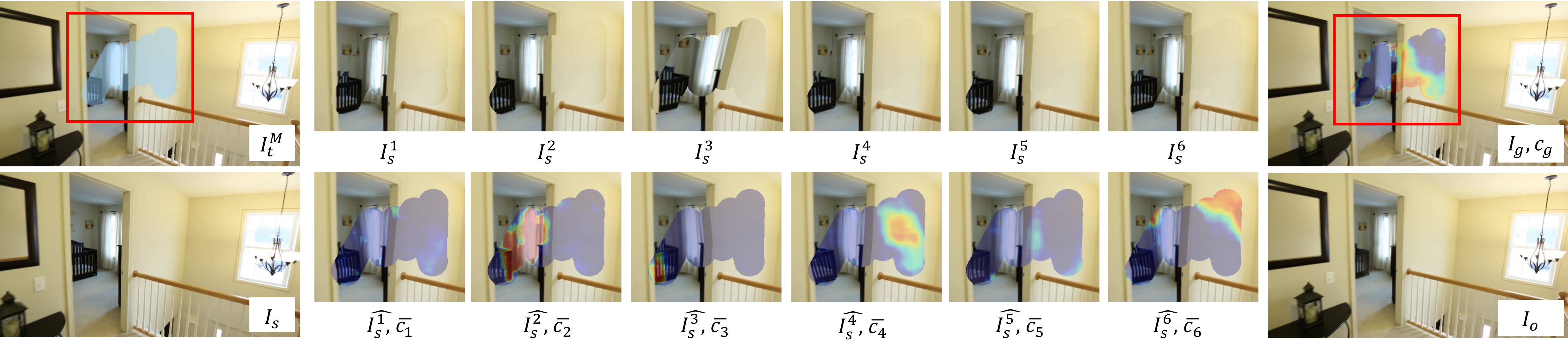}
\caption{Visualization of intermediate results. ${I_s^i}$ are the initialized homography warping of the source image $I_s$. $\hat{I_s^i}$ are the learned spatial and color transformation of ${I_s^i}$. The final result $I_o$ is the merging of $\hat{I_s^i}$ and the results of ProFill $I_g$ by pixel-wise weights $\overline{c_i}$ and $c_g$ overlaying on the images. The result draws from regions that are better aligned: on the left  from $\hat{I_s^2}$, in the center from the single image inpainting $I_g$, and on the right from $\hat{I_s^4}$ and $\hat{I_s^6}$. Zoom in for better visualization.} 
  \label{fig:res2}
\end{figure*}

\textbf{Results on User-Provided Images.}
In Figure \ref{fig:res1}, we show visual results of testing on real user-provided images. We indicate the hole region on the target image, and crop only the region of interest due to the space limits. More results can be found in the appendix. APAP and DFG well-preserve the source image contents due to the global homography warping, but they still suffer from color inconsistencies and alignment issues. We also experimented with combining Poisson blending with APAP but found it gives color bleeding artifacts: see the appendix for details. OPN usually works well when there are multiple reference frames which have similar scales and color distributions within the same video clips. However, if only one source reference image exists, the non-local attention module struggles to search for similar local patches and fails. ProFill with the contextual attention module usually does well in searching for textures, but the estimated intermediate coarse results cannot be matched with specific image contents. Thus the reference-based ProFill can only achieve texture or object removal but not background contents recovery. Compared to them, ours better reuses the background patterns and achieves a content-aware alignment and composition. The generated results are more faithful to and compatible with the target image. The multi-homography proposal approach provides more options for warping initialization when the matched features are too complex for a single homography. It helps to resolve challenging cases when the hole regions do not belong to the dominant plane in the image as shown in row four. In Figure \ref{fig:res2} we show intermediate results indicating regions selected to form the final results and how the CST resolves misalignment. 

\subsection{Quantitative Comparison}

\textbf{Results on RealEstate10K.}
The quantitative comparison on RealEstate10K is shown in Table \ref{exp:quant}. OPN and ProFill are more suitable for large batch testing. We tested them on the entire testing set. 
Results on cropped image pairs with Frame Displacement (FD) 10, 20 and 30 are reported in terms of PSNR, SSIM and LPIPS scores \cite{zhang2018perceptual} based on AlexNet \cite{krizhevsky2017imagenet}. APAP and DFG are not suitable for large batch testing and their performance may be influenced by non-existing regions, so we sampled a 300-image subset from FD=10 as \textit{Small Set} to test. Results showed that contextual-attention based ProFill failed to faithfully reconstruct the source contents. Optical-flow based DFG achieved better results by smoothly completing the flow field. OPN with atomic patch matching was not better than our warping-based approach. The TransFill thus demonstrated its superiority in faithful reconstruction. 

\textbf{User Study on User-Provided Images.}
To better evaluate the performance on our user-provided images, we conducted a user study via Amazon Mechanical Turk (AMT). We compared our method with each baseline separately and presented users with binary choice questions. We requested the users to choose one fill result which looks more realistic and faithful. 
To ensure the reliability, we used a pre-qualification test as well as check questions, as we explain in the appendix. For each method pair, we randomly sampled 80 examples, and each example was evaluated by 7 independent users. For each sample, one method was regarded as ``preferred" if at least 5 users selected it. Samples voted by 3 or 4 users are considered confusing samples and filtered out. We reported TransFill's Preference Rate (PR) in Table \ref{exp:quant}. The high preference rate demonstrates the effectiveness of TransFill. We also conducted a one-sample permutation t-test with $10^6$ samples by assuming a null hypothesis that on average 3.5 users prefer one method. The p-values are all sufficiently small so we can draw the conclusion that the preference for our method was statistically significant.  
\begin{table*}[t]
\setlength{\abovecaptionskip}{0pt}
\centering
\footnotesize
\caption{Quantitative Comparisons and User Study. \textbf{FD}: Frame Displacement. \textbf{PR}: Preference Rate}
\resizebox{\textwidth}{!}{
\begin{tabular}{|r||c|c|c|c|c||c|c|}
\hline
&\multicolumn{5}{c||}{RealEstate10K: PSNR $\uparrow$/ SSIM $\uparrow$  / LPIPS $\downarrow$ }&\multicolumn{2}{c|}{User-provided Images: User Study}\\\hline
Model&FD=10&FD=20&FD=30 &All&Small Set&PR & p-value \\ \hline\hline
APAP \cite{zaragoza2013projective}& -&-&- &-&31.94 / 0.9738 / 0.0251 &90.76$\%$ &$p<10^{-6}$\\\hline
DFG \cite{xu2019deep}&- &- &- &- & 36.17 / 0.9873 / 0.0155 &87.50$\%$  &$p<10^{-6}$  \\\hline
OPN \cite{oh2019onion}&33.45 / 0.9765 / 0.0201 &32.47 / 0.9734 / 0.0258 &31.32 / 0.9699 / 0.0320 &32.43 / 0.9734 / 0.0261 &33.40 / 0.9771 / 0.0207 &95.65$\%$ &$p<10^{-6}$ \\\hline
ProFill~\cite{zeng2020high}&31.18 / 0.9689 / 0.0423 &31.14 / 0.9687 / 0.0425 &30.83 / 0.9683 / 0.0440 & 31.05 / 0.9687 / 0.0429 &30.95 / 0.9690 / 0.0419 &81.67$\%$  & $p<10^{-6}$\\\hline
TransFill (Ours)&\bf{39.59 / 0.9919 / 0.0116} &\bf{37.39 / 0.9877 / 0.0162} &\bf{35.62 / 0.9839 / 0.0213}& \bf{37.58 / 0.9879 / 0.0164}& \bf{38.83 / 0.9914 / 0.0126}&- &- \\
\hline
\end{tabular}}
\label{exp:quant}
\vspace{-4mm}
\end{table*}

\subsection{Ablation Study} 
\textbf{Type and Number of Multi-Homography Proposals.} 
This ablation study was conducted on the testing set of the RealEstate10K. For each alternative, we re-trained the model. We compared the proposed depth-based points clustering methods with other alternatives including random and spatial clustering in Table \ref{exp:abla1}. When we proposed five homography matrices, depth-based clustering works best. The results were fairly close when we set $N$ to either 3 or 5, but $N=5$ was slightly better in PSNR. However, these are much  better than using just one global homography. 

\begin{table}[t]\setlength{\tabcolsep}{12pt}
\setlength{\abovecaptionskip}{0pt}
\centering
\footnotesize
\caption{Ablation Study on Multi-Homography Proposals. }
\resizebox{\columnwidth}{!}{
\begin{tabular}{|r|c|c|c|c|c|}
\hline
Clustering & N & Outlier Rejection & PSNR$\uparrow$ &SSIM$\uparrow$ &LPIPS$\downarrow$\\ \hline\hline
Depth &N=5 & OANet &\bf{37.576}&\bf{0.9879}&0.0164\\\hline
Depth &N=5&Ratio Test~\cite{lowe2004distinctive}&37.444&0.9876&0.0168\\\hline
Random&N=5& OANet &37.499&0.9873&0.0166\\\hline
Spatial&N=5&  OANet &37.384&0.9876&0.0169\\\hline
Depth &N=3&  OANet &37.537&0.9878&\bf{0.0162}\\\hline
None &N=1&  OANet &37.092&0.9868&0.0172\\\hline
\end{tabular}}
\label{exp:abla1}
\vspace{-4mm}
\end{table}

\textbf{Color-Spatial Transformation Module.}
Table \ref{exp:abla2} shows that the order of the Color-Spatial Transformer did not make too much difference. However, according to our experiments, adjusting the color first made the training converge faster since the guidance map was computed from a fixed $I_s^i$. Table \ref{exp:abla2} also demonstrated that both the Color and Spatial Transformer were necessary. 
\begin{table}[t]\setlength{\tabcolsep}{17pt}
\setlength{\abovecaptionskip}{0pt}
\centering
\footnotesize
\caption{Color-Spatial Transformation. \textbf{C}: Color, \textbf{S}:Spatial}
\resizebox{\columnwidth}{!}{
\begin{tabular}{|r|c|c|c|}
\hline
Order&PSNR$\uparrow$&SSIM$\uparrow$&LPIPS$\downarrow$\\ \hline\hline
$C\rightarrow S$&\bf{37.576}&\bf{0.9879}&0.0164\\\hline
$S\rightarrow C$&37.566 &\bf{0.9879} &\bf{0.0163}\\\hline
Only S&36.717&0.9866&0.0182\\\hline
Only C&36.228&0.9849&0.0179\\\hline
\end{tabular}}
\label{exp:abla2}
\vspace{-4mm}
\end{table}

\textbf{Pipeline Components.}
Table \ref{exp:abla3} indicates that refining the source image with CST outperforms directly merging the initialized homography-warped images. SPF and its output confidence $c_i$ effectively guided the learning of MPF. The proposed full pipeline achieved the best performance. 

\begin{table}[t]\setlength{\tabcolsep}{13pt}
\setlength{\abovecaptionskip}{0pt}
\centering
\footnotesize
\caption{Ablation Study on Pipeline Components. \textbf{CST}: Color-Spatial Transformer, \textbf{SPF}: Single-Proposal Fusion.}
\resizebox{\columnwidth}{!}{
\begin{tabular}{|c|c|c|c|c|}
\hline
CST&SPF&PSNR$\uparrow$&SSIM$\uparrow$&LPIPS$\downarrow$\\ \hline\hline
\checkmark&\checkmark&\bf{37.576}&\bf{0.9879}&\bf{0.0164}\\\hline
\xmark&\checkmark&35.579&0.9838&0.0183\\\hline
\checkmark&\xmark&36.710&0.9861&0.0188\\\hline
\xmark&\xmark&33.484&0.9782&0.0249\\
\hline
\end{tabular}}
\label{exp:abla3}
\vspace{-4mm}
\end{table}

\section{Limitations, Discussion and Conclusions}
The proposed method has limitations in certain situations. First,
the pipeline may not work well on extreme low-light or texture inputs containing very few SIFT feature points.  Second, our homography-based transformation is not suited for image pairs with extreme viewpoint changes.
Third, the current model may struggle to transfer color if the lighting environment is very different, such as day to night. This is because we use an effective bilateral grid color matching, but do not incorporate any specialized models that reason further about lighting  (e.g.~\cite{garon2019fast}). 
Additionally, we utilize the pre-trained ProFill to fill the missing pixels so 
the final generation quality highly depends on the performance of the single image inpainting module ProFill. That module could be replaced by other state-of-the-art models, and could potentially be optimized with the multi-fusion pipeline together. We leave that for future work.

In conclusion, we contribute a multi-source image inpainting model based on multiple homography, deep warping and color harmonization. The results outperform state-of-the-art single image and multi-source inpainting methods, especially when the hole contains complicated depth. 

{\small
\bibliographystyle{ieee_fullname}
\bibliography{egbib}

\begin{thebibliography}{10}\itemsep=-1pt

\bibitem{agarwala2004interactive}
Aseem Agarwala, Mira Dontcheva, Maneesh Agrawala, Steven Drucker, Alex Colburn,
  Brian Curless, David Salesin, and Michael Cohen.
\newblock Interactive digital photomontage.
\newblock In {\em ACM SIGGRAPH 2004 Papers}, pages 294--302. 2004.

\bibitem{baker2011database}
Simon Baker, Daniel Scharstein, JP Lewis, Stefan Roth, Michael~J Black, and
  Richard Szeliski.
\newblock A database and evaluation methodology for optical flow.
\newblock {\em International journal of computer vision}, 92(1):1--31, 2011.

\bibitem{ballester2001filling}
Coloma Ballester, Marcelo Bertalmio, Vicent Caselles, Guillermo Sapiro, and
  Joan Verdera.
\newblock Filling-in by joint interpolation of vector fields and gray levels.
\newblock {\em IEEE transactions on image processing}, 10(8):1200--1211, 2001.

\bibitem{barnes2009patchmatch}
Connelly Barnes, Eli Shechtman, Adam Finkelstein, and Dan~B Goldman.
\newblock Patchmatch: A randomized correspondence algorithm for structural
  image editing.
\newblock {\em ACM Trans. Graph.}, 28(3):24, 2009.

\bibitem{bertalmio2000image}
Marcelo Bertalmio, Guillermo Sapiro, Vincent Caselles, and Coloma Ballester.
\newblock Image inpainting.
\newblock In {\em Proceedings of the 27th annual conference on Computer
  graphics and interactive techniques}, pages 417--424, 2000.

\bibitem{fivek}
Vladimir Bychkovsky, Sylvain Paris, Eric Chan, and Fr{\'e}do Durand.
\newblock Learning photographic global tonal adjustment with a database of
  input / output image pairs.
\newblock In {\em The Twenty-Fourth IEEE Conference on Computer Vision and
  Pattern Recognition}, 2011.

\bibitem{cong2020dovenet}
Wenyan Cong, Jianfu Zhang, Li Niu, Liu Liu, Zhixin Ling, Weiyuan Li, and Liqing
  Zhang.
\newblock Dovenet: Deep image harmonization via domain verification.
\newblock In {\em Proceedings of the IEEE/CVF Conference on Computer Vision and
  Pattern Recognition}, pages 8394--8403, 2020.

\bibitem{detone2016deep}
Daniel DeTone, Tomasz Malisiewicz, and Andrew Rabinovich.
\newblock Deep image homography estimation.
\newblock {\em arXiv preprint arXiv:1606.03798}, 2016.

\bibitem{fischler1981random}
Martin~A Fischler and Robert~C Bolles.
\newblock Random sample consensus: a paradigm for model fitting with
  applications to image analysis and automated cartography.
\newblock {\em Communications of the ACM}, 24(6):381--395, 1981.

\bibitem{flynn2016deepstereo}
John Flynn, Ivan Neulander, James Philbin, and Noah Snavely.
\newblock Deepstereo: Learning to predict new views from the world's imagery.
\newblock In {\em Proceedings of the IEEE conference on computer vision and
  pattern recognition}, pages 5515--5524, 2016.

\bibitem{gao2011constructing}
Junhong Gao, Seon~Joo Kim, and Michael~S Brown.
\newblock Constructing image panoramas using dual-homography warping.
\newblock In {\em CVPR 2011}, pages 49--56. IEEE, 2011.

\bibitem{garon2019fast}
Mathieu Garon, Kalyan Sunkavalli, Sunil Hadap, Nathan Carr, and
  Jean-Fran{\c{c}}ois Lalonde.
\newblock Fast spatially-varying indoor lighting estimation.
\newblock In {\em Proceedings of the IEEE Conference on Computer Vision and
  Pattern Recognition}, pages 6908--6917, 2019.

\bibitem{gharbi2017deep}
Micha{\"e}l Gharbi, Jiawen Chen, Jonathan~T Barron, Samuel~W Hasinoff, and
  Fr{\'e}do Durand.
\newblock Deep bilateral learning for real-time image enhancement.
\newblock {\em ACM Transactions on Graphics (TOG)}, 36(4):1--12, 2017.

\bibitem{granados2012background}
Miguel Granados, Kwang~In Kim, James Tompkin, Jan Kautz, and Christian
  Theobalt.
\newblock Background inpainting for videos with dynamic objects and a
  free-moving camera.
\newblock In {\em European Conference on Computer Vision}, pages 682--695.
  Springer, 2012.

\bibitem{he2015delving}
Kaiming He, Xiangyu Zhang, Shaoqing Ren, and Jian Sun.
\newblock Delving deep into rectifiers: Surpassing human-level performance on
  imagenet classification.
\newblock In {\em Proceedings of the IEEE international conference on computer
  vision}, pages 1026--1034, 2015.

\bibitem{herrmann2018robust}
Charles Herrmann, Chen Wang, Richard~Strong Bowen, Emil Keyder, Michael
  Krainin, Ce Liu, and Ramin Zabih.
\newblock Robust image stitching with multiple registrations.
\newblock In {\em Proceedings of the European Conference on Computer Vision
  (ECCV)}, pages 53--67, 2018.

\bibitem{horn1993determining}
Berthold~KP Horn and Brian~G. Schunck.
\newblock " determining optical flow": A retrospective.
\newblock 1993.

\bibitem{Hu2019RevisitingSI}
Junjie Hu, Mete Ozay, Yan Zhang, and Takayuki Okatani.
\newblock Revisiting single image depth estimation: Toward higher resolution
  maps with accurate object boundaries.
\newblock 2019.

\bibitem{iizuka2017globally}
Satoshi Iizuka, Edgar Simo-Serra, and Hiroshi Ishikawa.
\newblock Globally and locally consistent image completion.
\newblock {\em ACM Transactions on Graphics (ToG)}, 36(4):1--14, 2017.

\bibitem{ilg2017flownet}
Eddy Ilg, Nikolaus Mayer, Tonmoy Saikia, Margret Keuper, Alexey Dosovitskiy,
  and Thomas Brox.
\newblock Flownet 2.0: Evolution of optical flow estimation with deep networks.
\newblock In {\em Proceedings of the IEEE conference on computer vision and
  pattern recognition}, pages 2462--2470, 2017.

\bibitem{jaderberg2015spatial}
Max Jaderberg, Karen Simonyan, Andrew Zisserman, et~al.
\newblock Spatial transformer networks.
\newblock In {\em Advances in neural information processing systems}, pages
  2017--2025, 2015.

\bibitem{jia2006drag}
Jiaya Jia, Jian Sun, Chi-Keung Tang, and Heung-Yeung Shum.
\newblock Drag-and-drop pasting.
\newblock {\em ACM Transactions on graphics (TOG)}, 25(3):631--637, 2006.

\bibitem{johnson1967hierarchical}
Stephen~C Johnson.
\newblock Hierarchical clustering schemes.
\newblock {\em Psychometrika}, 32(3):241--254, 1967.

\bibitem{kingma2014adam}
Diederik~P Kingma and Jimmy Ba.
\newblock Adam: A method for stochastic optimization.
\newblock {\em arXiv preprint arXiv:1412.6980}, 2014.

\bibitem{krizhevsky2017imagenet}
Alex Krizhevsky, Ilya Sutskever, and Geoffrey~E Hinton.
\newblock Imagenet classification with deep convolutional neural networks.
\newblock {\em Communications of the ACM}, 60(6):84--90, 2017.

\bibitem{lee2020warping}
Kyu-Yul Lee and Jae-Young Sim.
\newblock Warping residual based image stitching for large parallax.
\newblock In {\em Proceedings of the IEEE/CVF Conference on Computer Vision and
  Pattern Recognition}, pages 8198--8206, 2020.

\bibitem{lee2019copy}
Sungho Lee, Seoung~Wug Oh, DaeYeun Won, and Seon~Joo Kim.
\newblock Copy-and-paste networks for deep video inpainting.
\newblock In {\em Proceedings of the IEEE International Conference on Computer
  Vision}, pages 4413--4421, 2019.

\bibitem{li2015dual}
Shiwei Li, Lu Yuan, Jian Sun, and Long Quan.
\newblock Dual-feature warping-based motion model estimation.
\newblock In {\em Proceedings of the IEEE International Conference on Computer
  Vision}, pages 4283--4291, 2015.

\bibitem{liao2020guidance}
Liang Liao, Jing Xiao, Zheng Wang, Chia-wen Lin, and Shin'ichi Satoh.
\newblock Guidance and evaluation: Semantic-aware image inpainting for mixed
  scenes.
\newblock {\em arXiv preprint arXiv:2003.06877}, 2020.

\bibitem{liu2009content}
Feng Liu, Michael Gleicher, Hailin Jin, and Aseem Agarwala.
\newblock Content-preserving warps for 3d video stabilization.
\newblock {\em ACM Transactions on Graphics (TOG)}, 28(3):1--9, 2009.

\bibitem{liu2018image}
Guilin Liu, Fitsum~A Reda, Kevin~J Shih, Ting-Chun Wang, Andrew Tao, and Bryan
  Catanzaro.
\newblock Image inpainting for irregular holes using partial convolutions.
\newblock In {\em Proceedings of the European Conference on Computer Vision
  (ECCV)}, pages 85--100, 2018.

\bibitem{liu2016meshflow}
Shuaicheng Liu, Ping Tan, Lu Yuan, Jian Sun, and Bing Zeng.
\newblock Meshflow: Minimum latency online video stabilization.
\newblock In {\em European Conference on Computer Vision}, pages 800--815.
  Springer, 2016.

\bibitem{lowe1999object}
David~G Lowe.
\newblock Object recognition from local scale-invariant features.
\newblock In {\em Proceedings of the seventh IEEE international conference on
  computer vision}, volume~2, pages 1150--1157. Ieee, 1999.

\bibitem{lowe2004distinctive}
David~G Lowe.
\newblock Distinctive image features from scale-invariant keypoints.
\newblock {\em International journal of computer vision}, 60(2):91--110, 2004.

\bibitem{mildenhall2019local}
Ben Mildenhall, Pratul~P Srinivasan, Rodrigo Ortiz-Cayon, Nima~Khademi
  Kalantari, Ravi Ramamoorthi, Ren Ng, and Abhishek Kar.
\newblock Local light field fusion: Practical view synthesis with prescriptive
  sampling guidelines.
\newblock {\em ACM Transactions on Graphics (TOG)}, 38(4):1--14, 2019.

\bibitem{nazeri2019edgeconnect}
Kamyar Nazeri, Eric Ng, Tony Joseph, Faisal Qureshi, and Mehran Ebrahimi.
\newblock Edgeconnect: Generative image inpainting with adversarial edge
  learning.
\newblock 2019.

\bibitem{newson2014video}
Alasdair Newson, Andr{\'e}s Almansa, Matthieu Fradet, Yann Gousseau, and
  Patrick P{\'e}rez.
\newblock Video inpainting of complex scenes.
\newblock {\em Siam journal on imaging sciences}, 7(4):1993--2019, 2014.

\bibitem{nguyen2018unsupervised}
Ty Nguyen, Steven~W Chen, Shreyas~S Shivakumar, Camillo~Jose Taylor, and Vijay
  Kumar.
\newblock Unsupervised deep homography: A fast and robust homography estimation
  model.
\newblock {\em IEEE Robotics and Automation Letters}, 3(3):2346--2353, 2018.

\bibitem{oh2019onion}
Seoung~Wug Oh, Sungho Lee, Joon-Young Lee, and Seon~Joo Kim.
\newblock Onion-peel networks for deep video completion.
\newblock In {\em Proceedings of the IEEE International Conference on Computer
  Vision}, pages 4403--4412, 2019.

\bibitem{paszke2019pytorch}
Adam Paszke, Sam Gross, Francisco Massa, Adam Lerer, James Bradbury, Gregory
  Chanan, Trevor Killeen, Zeming Lin, Natalia Gimelshein, Luca Antiga, et~al.
\newblock Pytorch: An imperative style, high-performance deep learning library.
\newblock In {\em Advances in neural information processing systems}, pages
  8026--8037, 2019.

\bibitem{pathak2016context}
Deepak Pathak, Philipp Krahenbuhl, Jeff Donahue, Trevor Darrell, and Alexei~A
  Efros.
\newblock Context encoders: Feature learning by inpainting.
\newblock In {\em Proceedings of the IEEE conference on computer vision and
  pattern recognition}, pages 2536--2544, 2016.

\bibitem{perez2003poisson}
Patrick P{\'e}rez, Michel Gangnet, and Andrew Blake.
\newblock Poisson image editing.
\newblock In {\em ACM SIGGRAPH 2003 Papers}, pages 313--318. 2003.

\bibitem{pitie2005n}
Francois Pitie, Anil~C Kokaram, and Rozenn Dahyot.
\newblock N-dimensional probability density function transfer and its
  application to color transfer.
\newblock In {\em Tenth IEEE International Conference on Computer Vision
  (ICCV'05) Volume 1}, volume~2, pages 1434--1439. IEEE, 2005.

\bibitem{reinhard2001color}
Erik Reinhard, Michael Adhikhmin, Bruce Gooch, and Peter Shirley.
\newblock Color transfer between images.
\newblock {\em IEEE Computer graphics and applications}, 21(5):34--41, 2001.

\bibitem{ren2019structureflow}
Yurui Ren, Xiaoming Yu, Ruonan Zhang, Thomas~H. Li, Shan Liu, and Ge Li.
\newblock Structureflow: Image inpainting via structure-aware appearance flow.
\newblock In {\em IEEE International Conference on Computer Vision (ICCV)},
  2019.

\bibitem{ronneberger2015u}
Olaf Ronneberger, Philipp Fischer, and Thomas Brox.
\newblock U-net: Convolutional networks for biomedical image segmentation.
\newblock In {\em International Conference on Medical image computing and
  computer-assisted intervention}, pages 234--241. Springer, 2015.

\bibitem{sarlin2020superglue}
Paul-Edouard Sarlin, Daniel DeTone, Tomasz Malisiewicz, and Andrew Rabinovich.
\newblock Superglue: Learning feature matching with graph neural networks.
\newblock In {\em Proceedings of the IEEE/CVF Conference on Computer Vision and
  Pattern Recognition}, pages 4938--4947, 2020.

\bibitem{simonyan2014very}
Karen Simonyan and Andrew Zisserman.
\newblock Very deep convolutional networks for large-scale image recognition.
\newblock {\em arXiv preprint arXiv:1409.1556}, 2014.

\bibitem{song2018spg}
Yuhang Song, Chao Yang, Yeji Shen, Peng Wang, Qin Huang, and C-C~Jay Kuo.
\newblock Spg-net: Segmentation prediction and guidance network for image
  inpainting.
\newblock {\em arXiv preprint arXiv:1805.03356}, 2018.

\bibitem{sun2010secrets}
Deqing Sun, Stefan Roth, and Michael~J Black.
\newblock Secrets of optical flow estimation and their principles.
\newblock In {\em 2010 IEEE computer society conference on computer vision and
  pattern recognition}, pages 2432--2439. IEEE, 2010.

\bibitem{sun2018pwc}
Deqing Sun, Xiaodong Yang, Ming-Yu Liu, and Jan Kautz.
\newblock Pwc-net: Cnns for optical flow using pyramid, warping, and cost
  volume.
\newblock In {\em Proceedings of the IEEE conference on computer vision and
  pattern recognition}, pages 8934--8943, 2018.

\bibitem{sunkavalli2010multi}
Kalyan Sunkavalli, Micah~K Johnson, Wojciech Matusik, and Hanspeter Pfister.
\newblock Multi-scale image harmonization.
\newblock {\em ACM Transactions on Graphics (TOG)}, 29(4):1--10, 2010.

\bibitem{tao2010error}
Michael~W Tao, Micah~K Johnson, and Sylvain Paris.
\newblock Error-tolerant image compositing.
\newblock In {\em European Conference on Computer Vision}, pages 31--44.
  Springer, 2010.

\bibitem{tsai2017deep}
Yi-Hsuan Tsai, Xiaohui Shen, Zhe Lin, Kalyan Sunkavalli, Xin Lu, and Ming-Hsuan
  Yang.
\newblock Deep image harmonization.
\newblock In {\em Proceedings of the IEEE Conference on Computer Vision and
  Pattern Recognition}, pages 3789--3797, 2017.

\bibitem{tucker2020single}
Richard Tucker and Noah Snavely.
\newblock Single-view view synthesis with multiplane images.
\newblock In {\em Proceedings of the IEEE/CVF Conference on Computer Vision and
  Pattern Recognition}, pages 551--560, 2020.

\bibitem{wang2019underexposed}
Ruixing Wang, Qing Zhang, Chi-Wing Fu, Xiaoyong Shen, Wei-Shi Zheng, and Jiaya
  Jia.
\newblock Underexposed photo enhancement using deep illumination estimation.
\newblock In {\em Proceedings of the IEEE Conference on Computer Vision and
  Pattern Recognition}, pages 6849--6857, 2019.

\bibitem{weinzaepfel2013deepflow}
Philippe Weinzaepfel, Jerome Revaud, Zaid Harchaoui, and Cordelia Schmid.
\newblock Deepflow: Large displacement optical flow with deep matching.
\newblock In {\em Proceedings of the IEEE international conference on computer
  vision}, pages 1385--1392, 2013.

\bibitem{wexler2007space}
Yonatan Wexler, Eli Shechtman, and Michal Irani.
\newblock Space-time completion of video.
\newblock {\em IEEE Transactions on pattern analysis and machine intelligence},
  29(3):463--476, 2007.

\bibitem{WSZ09}
Oliver Whyte, Josef Sivic, and Andrew Zisserman.
\newblock Get out of my picture! internet-based inpainting.
\newblock In {\em Proceedings of the 20th British Machine Vision Conference,
  London}, 2009.

\bibitem{wulff2015efficient}
Jonas Wulff and Michael~J Black.
\newblock Efficient sparse-to-dense optical flow estimation using a learned
  basis and layers.
\newblock In {\em Proceedings of the IEEE Conference on Computer Vision and
  Pattern Recognition}, pages 120--130, 2015.

\bibitem{xiaodong2019improving}
Cun Xiaodong and Pun Chi-Man.
\newblock Improving the harmony of the composite image by spatial-separated
  attention module.
\newblock {\em arXiv preprint arXiv:1907.06406}, 2019.

\bibitem{xiong2019foreground}
Wei Xiong, Jiahui Yu, Zhe Lin, Jimei Yang, Xin Lu, Connelly Barnes, and Jiebo
  Luo.
\newblock Foreground-aware image inpainting.
\newblock In {\em Proceedings of the IEEE conference on computer vision and
  pattern recognition}, pages 5840--5848, 2019.

\bibitem{xu2019deep}
Rui Xu, Xiaoxiao Li, Bolei Zhou, and Chen~Change Loy.
\newblock Deep flow-guided video inpainting.
\newblock In {\em Proceedings of the IEEE Conference on Computer Vision and
  Pattern Recognition}, pages 3723--3732, 2019.

\bibitem{xue2015computational}
Tianfan Xue, Michael Rubinstein, Ce Liu, and William~T Freeman.
\newblock A computational approach for obstruction-free photography.
\newblock {\em ACM Transactions on Graphics (TOG)}, 34(4):1--11, 2015.

\bibitem{yang2017high}
Chao Yang, Xin Lu, Zhe Lin, Eli Shechtman, Oliver Wang, and Hao Li.
\newblock High-resolution image inpainting using multi-scale neural patch
  synthesis.
\newblock In {\em Proceedings of the IEEE Conference on Computer Vision and
  Pattern Recognition}, pages 6721--6729, 2017.

\bibitem{ye2019deepmeshflow}
Nianjin Ye, Chuan Wang, Shuaicheng Liu, Lanpeng Jia, Jue Wang, and Yongqing
  Cui.
\newblock Deepmeshflow: Content adaptive mesh deformation for robust image
  registration.
\newblock {\em arXiv preprint arXiv:1912.05131}, 2019.

\bibitem{yi2020contextual}
Zili Yi, Qiang Tang, Shekoofeh Azizi, Daesik Jang, and Zhan Xu.
\newblock Contextual residual aggregation for ultra high-resolution image
  inpainting.
\newblock In {\em Proceedings of the IEEE/CVF Conference on Computer Vision and
  Pattern Recognition}, pages 7508--7517, 2020.

\bibitem{yu2018generative}
Jiahui Yu, Zhe Lin, Jimei Yang, Xiaohui Shen, Xin Lu, and Thomas~S Huang.
\newblock Generative image inpainting with contextual attention.
\newblock In {\em Proceedings of the IEEE conference on computer vision and
  pattern recognition}, pages 5505--5514, 2018.

\bibitem{yu2019free}
Jiahui Yu, Zhe Lin, Jimei Yang, Xiaohui Shen, Xin Lu, and Thomas~S Huang.
\newblock Free-form image inpainting with gated convolution.
\newblock In {\em Proceedings of the IEEE International Conference on Computer
  Vision}, pages 4471--4480, 2019.

\bibitem{zaragoza2013projective}
Julio Zaragoza, Tat-Jun Chin, Michael~S Brown, and David Suter.
\newblock As-projective-as-possible image stitching with moving dlt.
\newblock In {\em Proceedings of the IEEE conference on computer vision and
  pattern recognition}, pages 2339--2346, 2013.

\bibitem{zeng2020high}
Yu Zeng, Zhe Lin, Jimei Yang, Jianming Zhang, Eli Shechtman, and Huchuan Lu.
\newblock High-resolution image inpainting with iterative confidence feedback
  and guided upsampling.
\newblock {\em arXiv preprint arXiv:2005.11742}, 2020.

\bibitem{zhang2019oanet}
Jiahui Zhang, Dawei Sun, Zixin Luo, Anbang Yao, Lei Zhou, Tianwei Shen, Yurong
  Chen, Long Quan, and Hongen Liao.
\newblock Learning two-view correspondences and geometry using order-aware
  network.
\newblock {\em International Conference on Computer Vision (ICCV)}, 2019.

\bibitem{zhang2019content}
Jirong Zhang, Chuan Wang, Shuaicheng Liu, Lanpeng Jia, Nianjin Ye, Jue Wang, Ji
  Zhou, and Jian Sun.
\newblock Content-aware unsupervised deep homography estimation.
\newblock {\em arXiv preprint arXiv:1909.05983}, 2019.

\bibitem{zhang2018perceptual}
Richard Zhang, Phillip Isola, Alexei~A Efros, Eli Shechtman, and Oliver Wang.
\newblock The unreasonable effectiveness of deep features as a perceptual
  metric.
\newblock In {\em CVPR}, 2018.

\bibitem{zhao2019guided}
Yinan Zhao, Brian Price, Scott Cohen, and Danna Gurari.
\newblock Guided image inpainting: Replacing an image region by pulling content
  from another image.
\newblock In {\em 2019 IEEE Winter Conference on Applications of Computer
  Vision (WACV)}, pages 1514--1523. IEEE, 2019.

\bibitem{zheng2019pluralistic}
Chuanxia Zheng, Tat-Jen Cham, and Jianfei Cai.
\newblock Pluralistic image completion.
\newblock In {\em Proceedings of the IEEE Conference on Computer Vision and
  Pattern Recognition}, pages 1438--1447, 2019.

\bibitem{zhou2018stereo}
Tinghui Zhou, Richard Tucker, John Flynn, Graham Fyffe, and Noah Snavely.
\newblock Stereo magnification: Learning view synthesis using multiplane
  images.
\newblock {\em arXiv preprint arXiv:1805.09817}, 2018.

\bibitem{zhu2015learning}
Jun-Yan Zhu, Philipp Krahenbuhl, Eli Shechtman, and Alexei~A Efros.
\newblock Learning a discriminative model for the perception of realism in
  composite images.
\newblock In {\em Proceedings of the IEEE International Conference on Computer
  Vision}, pages 3943--3951, 2015.

\end{thebibliography}
}

\clearpage

\appendix
\section{Appendix A: More Ablation Studies}
\renewcommand\thefigure{\thesection.\arabic{figure}}    
\setcounter{figure}{0} 

\subsection{Color-Spatial Transformer}
\begin{figure}[h]
\centering
  \includegraphics[width=\linewidth]{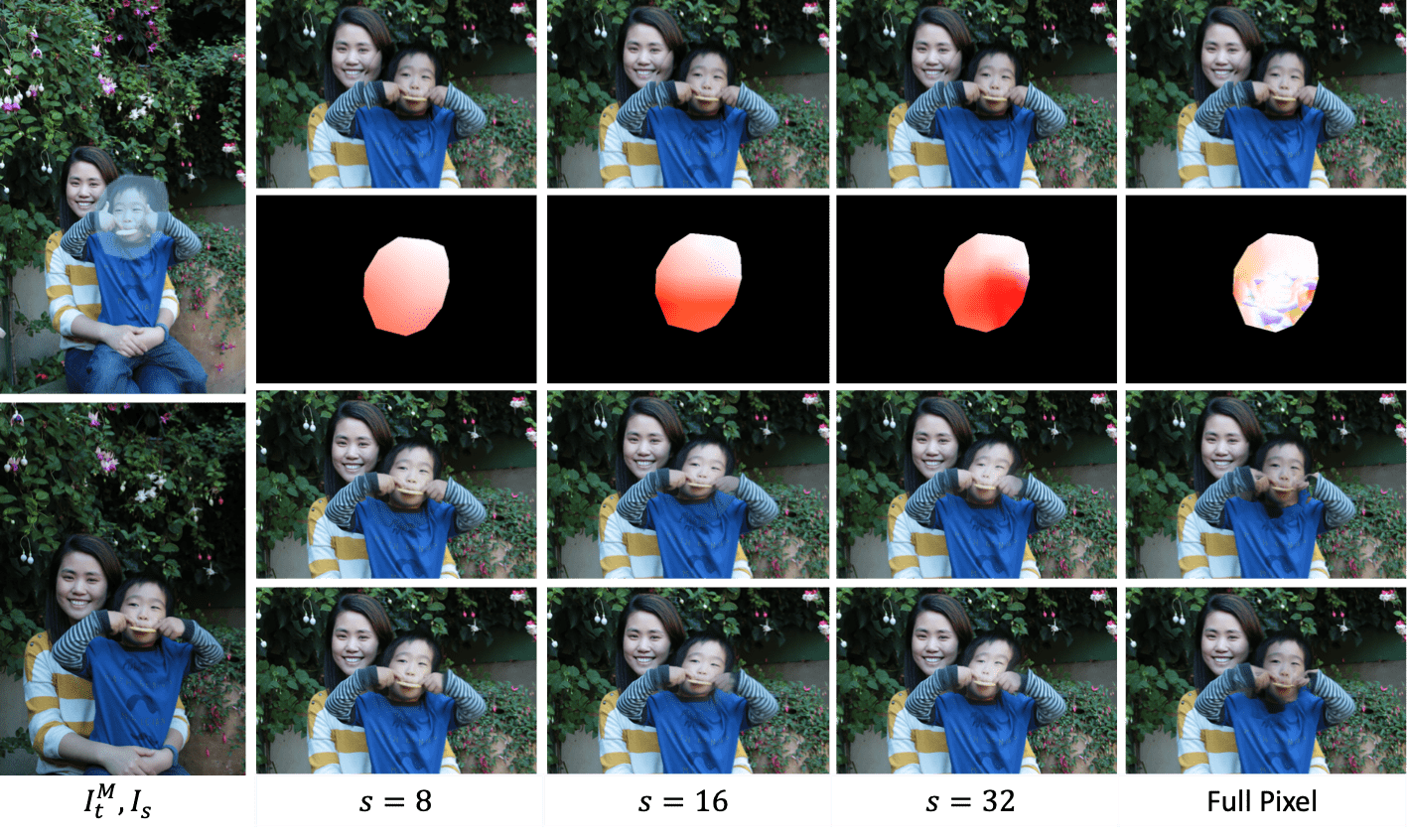}
\caption{Ablation study on the resolution setting in the Color-Spatial Transformer. First row, direct composition of target image $I_t^M$ and one of the single-homography transformed source images $I_s^i$. Second row, the learned pixel-wise warping field $A_s^i$ visualized using color wheel in \cite{baker2011database}. Third row, the color-spatial transformed image $\hat{I_s^i}$. Last row, the final merging result $I_o$.   }
  \label{fig:abla_s}
\end{figure}

Recall that while introducing the Color-Spatial Transformer, we intend to preserve the texture details and the rigidity of the source image contents. Therefore, given $A_c^i = [K_c^i \quad b_c^i] \in {\rm I\!R}^{W\times H \times 3 \times 4}$, and $\bar{A_s^i} = B_s(u_s^i) \in \rm I\!R^{s \times s \times 2}$, we fix $s = 8$ and $d = 8$ in our experiments. We find $d$ does not influence the performance a lot, and the guidance map is automatically learned to uniformly span the necessary bins like in the HDRNet\cite{gharbi2017deep}. Figure \ref{fig:abla_s} shows the comparison when we set different $s$ values. It suggests that increasing $s$ gives more degrees of freedom to the learned warping field $A_s^i$. However, while encountering larger holes like in Figure \ref{fig:abla_s}, better flexibility does not better align the contents as expected, but distorts the contents inside the hole. The transformed color field also becomes less smooth as $s$ increases. In an extreme case, suppose we replace the deep bilateral grid and directly learn a full-resolution pixel-wise color-warping field with total variance constraints as in the last column, the model struggles to infer a reasonable color-warping operation within a large hole. 

We conclude that CST with smaller $s$ value like $s=8$ generalizes better to inference images with varying spatial resolutions. It is mainly due to the ill-posedness of image completion. Unlike conventional image registration tasks where all the pixels of the matched regions are available, hole regions are missing in the inpainting task. Less freedom in the hole area preserves better content integrity and semantics.

\subsection{Network Components}
\begin{figure}[h]
\centering
  \includegraphics[width=\linewidth]{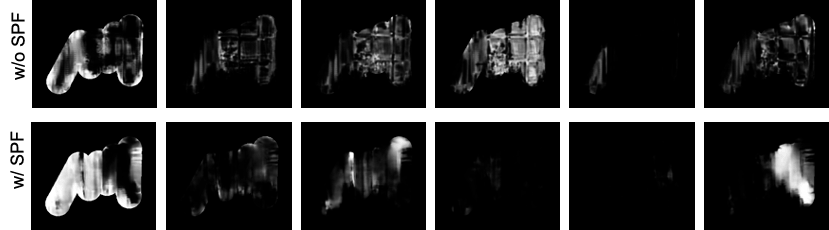}
\caption{Final fusion masks $\bar{c_i}$ learned by the model with or without the Single-Proposal Fusion (SPF) module. By using SPF outputs as guidance to learn the MPF, the final weights learned tend to be more sparse.}
  \label{fig:spfa}
  \vspace{-3mm}
\end{figure}
\textbf{Importance of Single-Proposal Fusion (SPF)}
Our experiments exhibit that the proposed Single-Proposal Fusion (SPF) module before the Multi-Proposal Fusion (MPF) is necessary for effectively learning the final merging weights of all the proposals. We find directly learning the weights to fuse all the proposals is very challenging. The learned weights have a hard time becoming sparse even though the same total variance loss is imposed. A comparison of the merging mask $\bar{c_i}$ between the model with and without SPF is shown in Figure \ref{fig:spfa}. Using SPF outputs $c_i$ as a structure guidance for learning the fusion of multiple proposals works better in practice. 
\begin{figure}[h]
\centering
  \includegraphics[width=\linewidth]{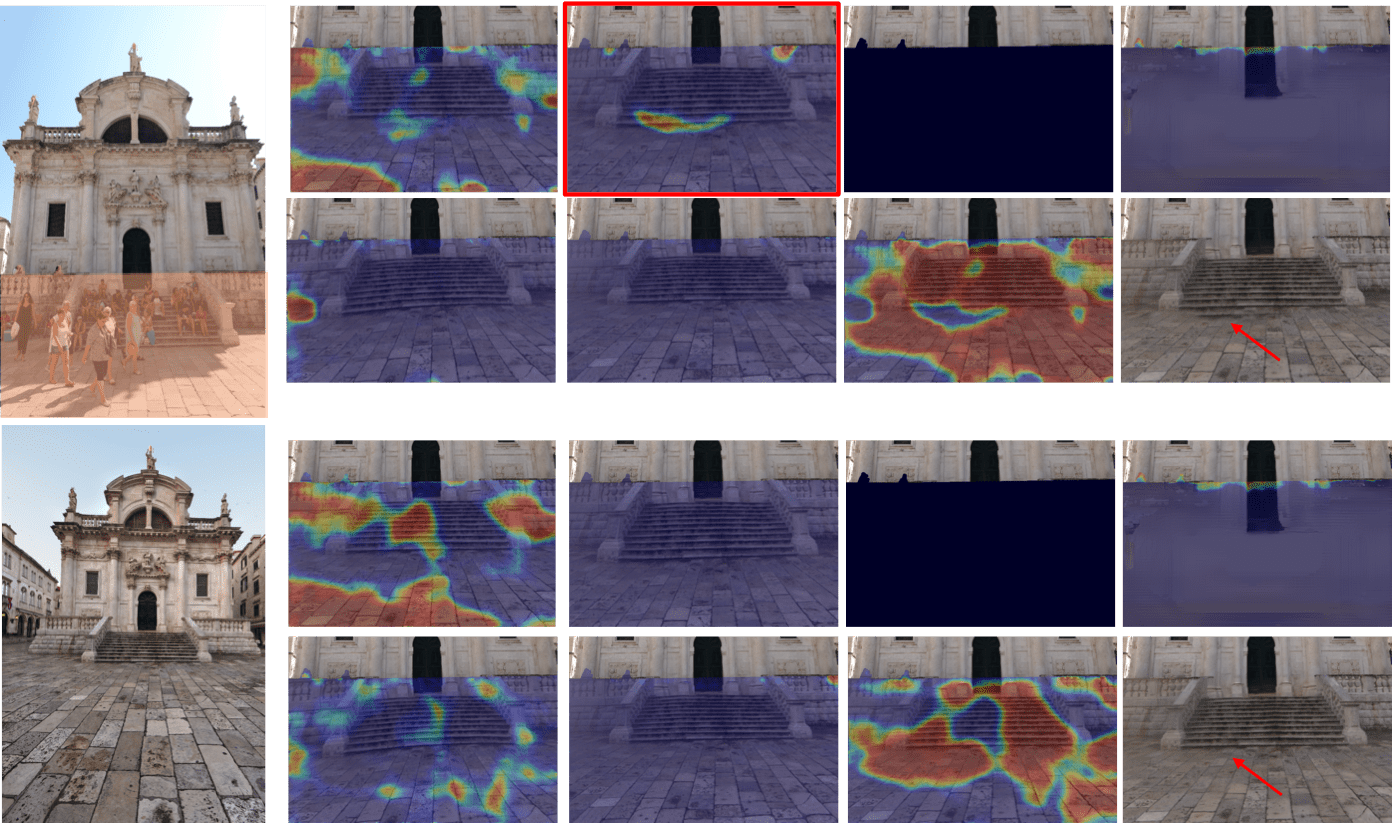}
\caption{One example of user-involved interactive editing. For the given target and source images on the left, we generate seven proposals including one $I_g$ from ProFill. For the upper group of images, we visualize the regions selected by our model to synthesize the final results. The image with red bounding box yields an unexpected artifacts of stairs. By zeroing out its corresponding $c_2$, we can correspondingly obtain zero-valued $\bar{c_2}$ as shown in the lower image group. Other maps are also correspondingly redistributed. The final result on the lower-right position is then generated by merging the other selected proposals with nonzero weighted masks. As we can see, the artifact disappears. }
  \label{fig:mask}
\end{figure}

\textbf{Correlation between $c_i$ and $\bar{c_i}$}
In our experiments, the learned single-proposal fusion mask $c_i$ and multi-proposal fusion mask $\bar{c_i}$ demonstrate strong correlation. Specifically, by zeroing out one of the $c_i$, the values in $\bar{c_i}$ will also vanish. This shows the MPF constructs the correspondence to make $\bar{c_i}$ be conditioned on $c_i$. This provides more flexibility for our model to incorporate user interactions. Suppose users want to eliminate the elements in some proposals, one can simply zero them out and the final results will only be merged from other selected proposals. Such a process is demonstrated in Figure \ref{fig:mask}. 

\begin{figure}[h]
\centering
  \includegraphics[width=\linewidth]{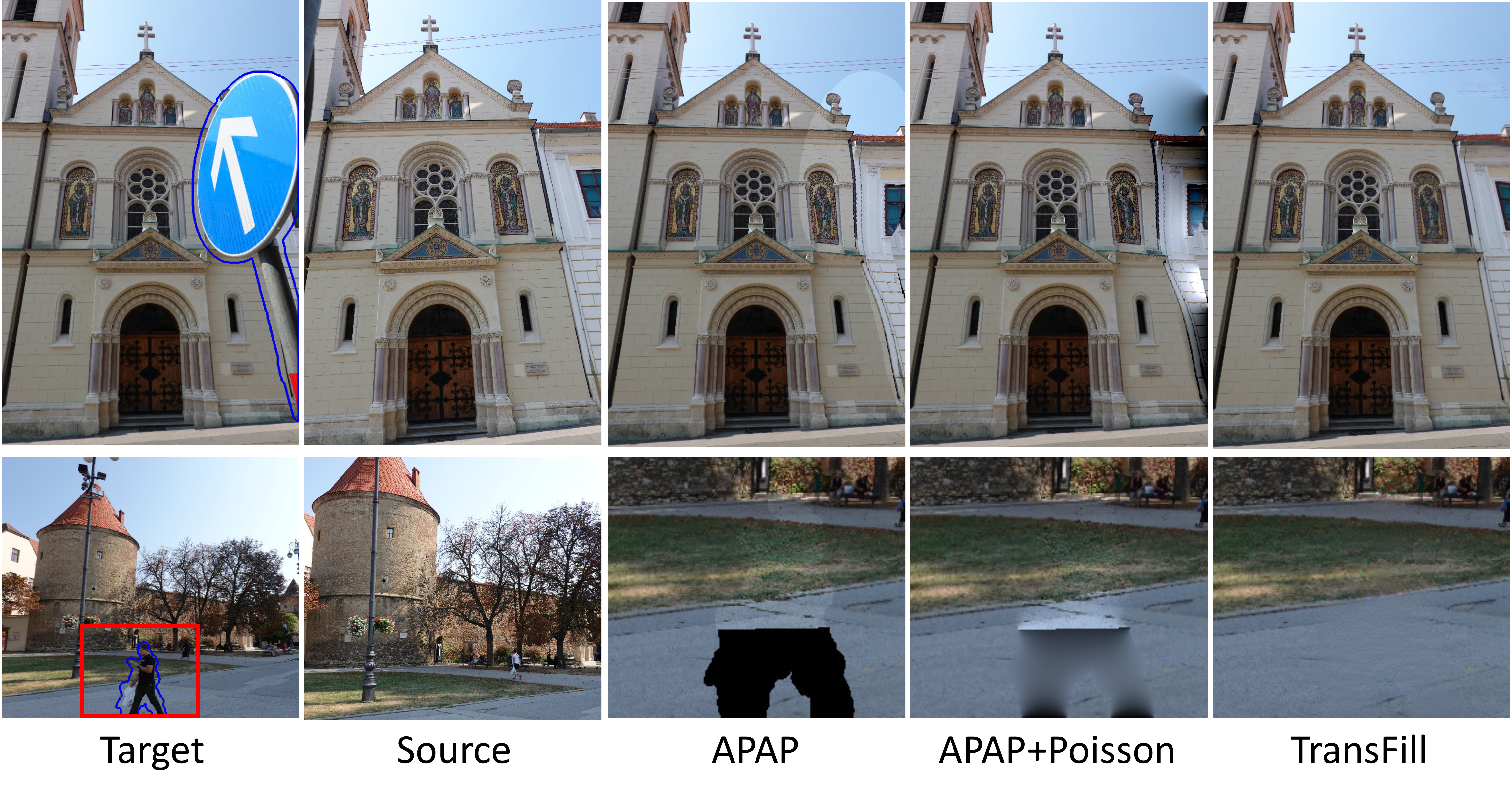}
\caption{Ablation study on APAP with Poisson blending post-processing. The color bleeding artifacts are significant in some cases when there are strong color mismatches or non-existing regions. APAP does not include image inpainting, so regions that are outside the source image appear as black.}
  \label{fig:poi}
\end{figure}
\subsection{APAP with Poisson Blending}
We experimented with using Poisson blending \cite{perez2003poisson} combined with APAP. The testing result on the \textit{Small Set} of images with only few non-existing regions is increased from 31.94dB / 0.9738 to 32.56dB / 0.9754 in terms of PSNR / SSIM. However, we did not incorporate Poisson blending in the baseline because we found in some cases there could be significant color bleeding artifacts due to strong color mismatches and non-existing regions especially along the boundary of the hole. Some visual comparisons are shown in Figure \ref{fig:poi}.

\begin{figure}[h]
\centering
  \includegraphics[width=\linewidth]{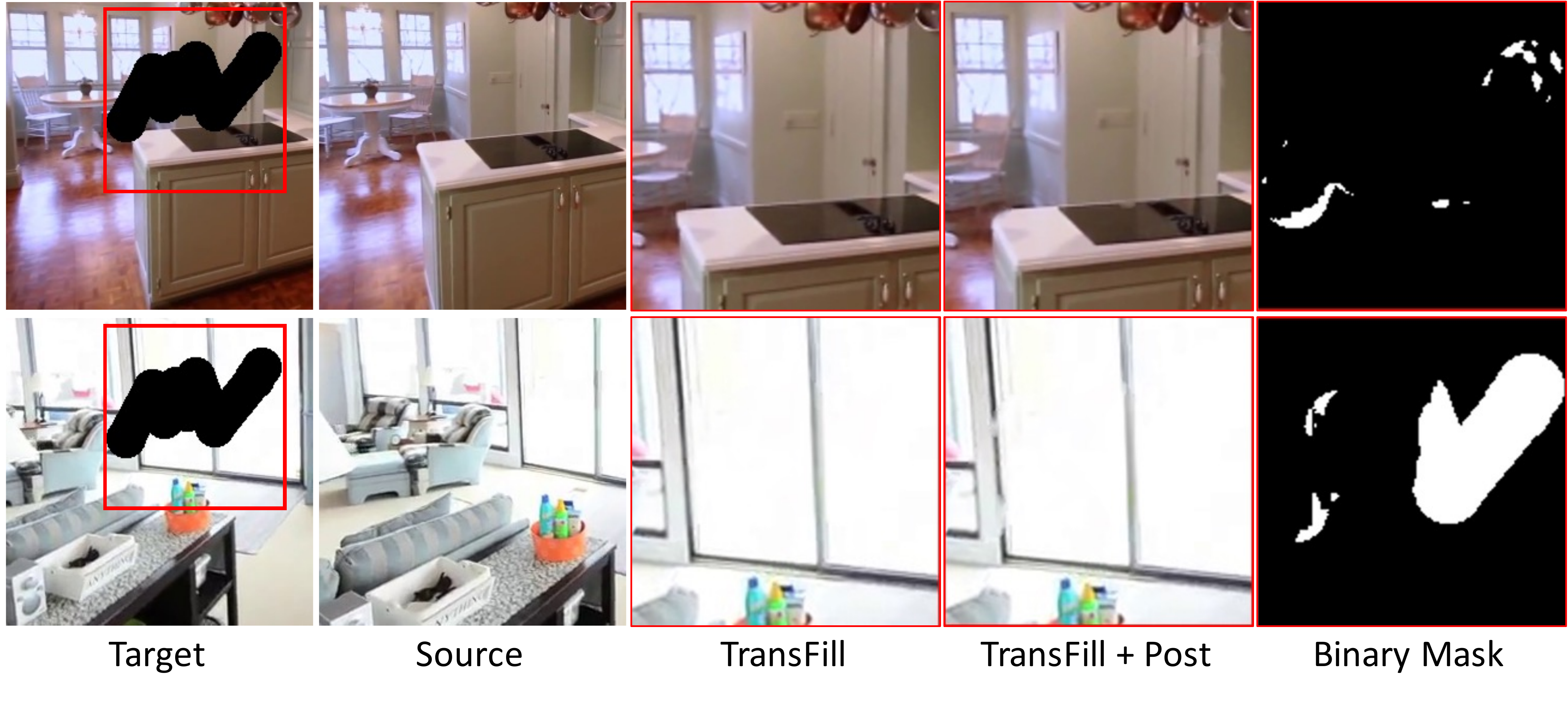}
\caption{Post-hoc refilling results using ProFill. The TransFill columns show a zoom of the original output, the post-hoc filling result (TransFill + Post), and the region to be refilled from the confidence $c_g$ (Binary Mask). The re-filling with a partial mask may introduce additional artifacts like broken door frames.  }
  \label{fig:ph}
\end{figure}
\subsection{Using ProFill with Partial Masks}
As we stated that single image techniques don’t work well for larger holes, while in our work, the single-image inpainting is computed over the full mask area. We also thought about using ProFill or other single-image inpainting method with partial mask, but could not find a principled and an end-to-end way to do this. However, we analyzed an approximation of this approach where we used the confidence map $c_g$ estimated by our method, and binarized it to do a post-hoc fill (with ProFill) of each hole region of the target that corresponds to single image inpainting (where the content is not visible in the source image or not well reused). Comparisons are shown in Figure \ref{fig:ph}. This reveals that since the mask was learned for merging purposes, a post-hoc filling using the mask may introduce other artifacts like broken door frames. The average testing results on RealEstate10K decreased from 37.58dB / 0.9879 / 0.0164 to 37.13dB / 0.9871 / 0.0173 in terms of PSNR / SSIM / LPIPS. However, using partial masks to fill only non-existing regions might work better for images with larger non-existing regions, and become more robust if another approach of learning is taken. 

\section{Appendix B: User Study Details}
\begin{figure}[h]
\centering
  \includegraphics[width=\linewidth]{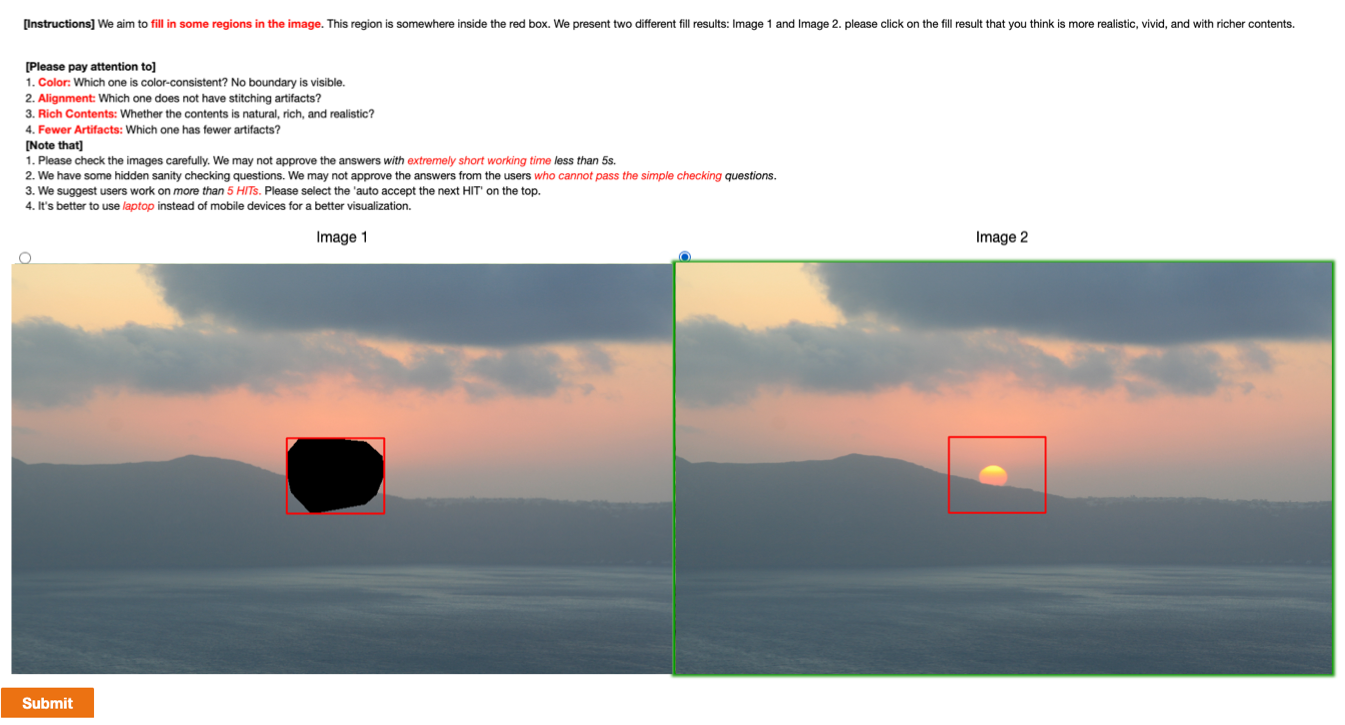}
\caption{User study interface and check questions. We set up 10 trivial questions to let the users choose which one is more completed.}
  \label{fig:us}
\end{figure}
The GUI of our user study at AMT is shown in Figure \ref{fig:us}. To guarantee the reliability of the users' feedback, we require the users to take a qualification test before they evaluate. The test presents users with the 10 trivial pairs $I_t$ and $I_t^M$ and users who answer correctly more than 8 questions are approved to take the official test. We also mix 10 random sanity check questions with the real questions. No users had to be disqualified due to failing the initial test, and only very few users (4 users) got check questions later in the study wrong (5.7\% of total opinions), so we conclude that the user responses are reliable. 

\section{Appendix C: Failure Cases}
\begin{figure}[h]
\centering
  \includegraphics[width=\linewidth]{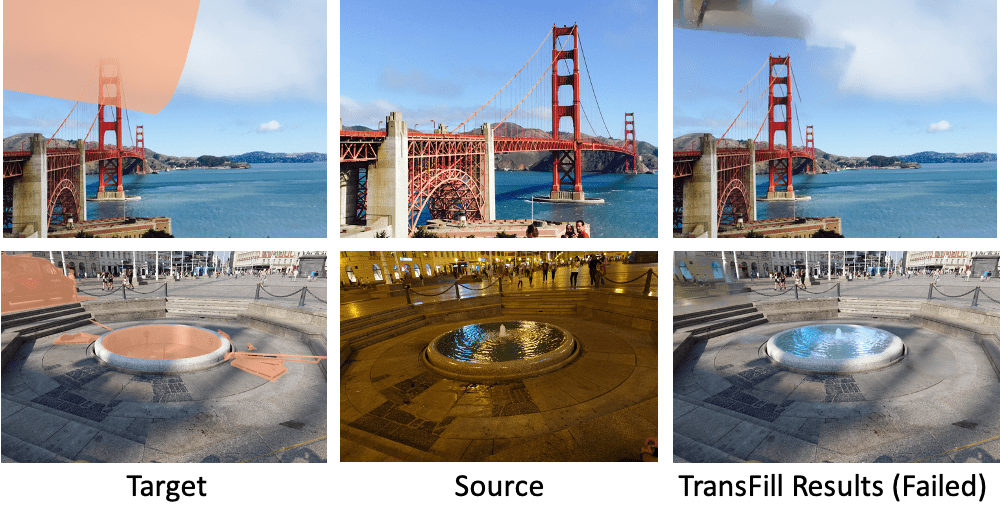}
\caption{Failure cases. These demonstrate limitations with large changes in viewing angle, outpainting artifacts from the off-the-shelf single image inpainting module ProFill, and challenges in handling dramatically different lighting environments.}
  \label{fig:fail}
\end{figure}
Figure \ref{fig:fail} shows some examples of failure cases when the viewing angle changes are large. The color matching module may struggle if there are extreme lighting differences. We may also encounter outpainting artifact issues caused by ProFill. 

\section{More Visual Results Comparison}
\begin{figure*}[th]
\centering
  \includegraphics[width=\linewidth]{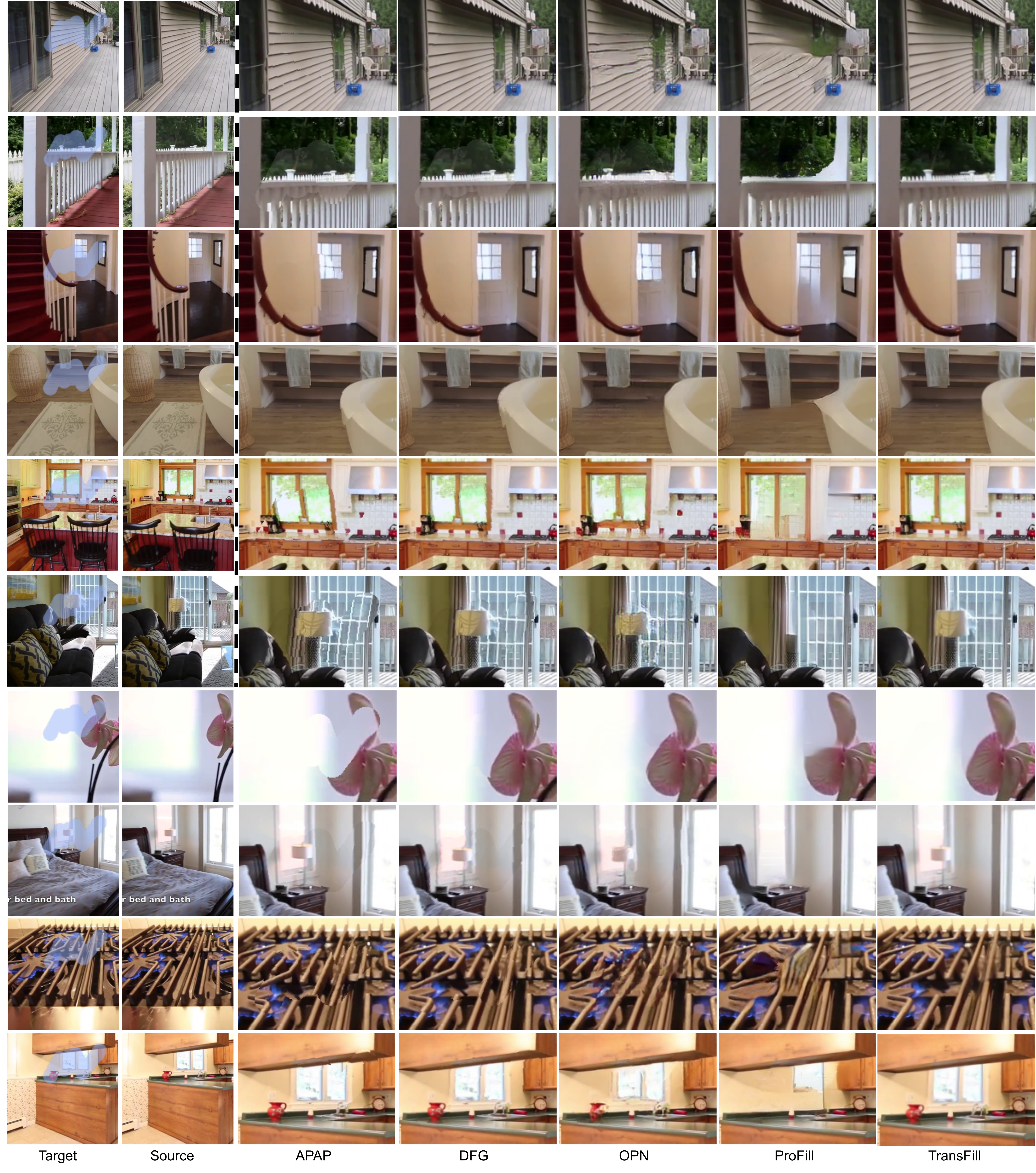}
\caption{Visual results comparison on the RealEstate10K dataset with FD=10. These have been cropped. \textbf{Please zoom in so that there are about 3-4 images across the width of the screen to reveal the significant differences in fine details.} Compared with the baselines, our proposed TransFill achieves better spatial alignment and faithfulness to the source image content.}
  \label{fig:real1}
\end{figure*}
\textbf{Visual Results on RealEstate10K}
We present more visual results in Figure \ref{fig:real1} on the RealEstate10K dataset. Compared with the baselines, our proposed TransFill achieves better spatial alignment and content faithfulness. 

\begin{figure*}[t]
\centering
  \includegraphics[width=\linewidth]{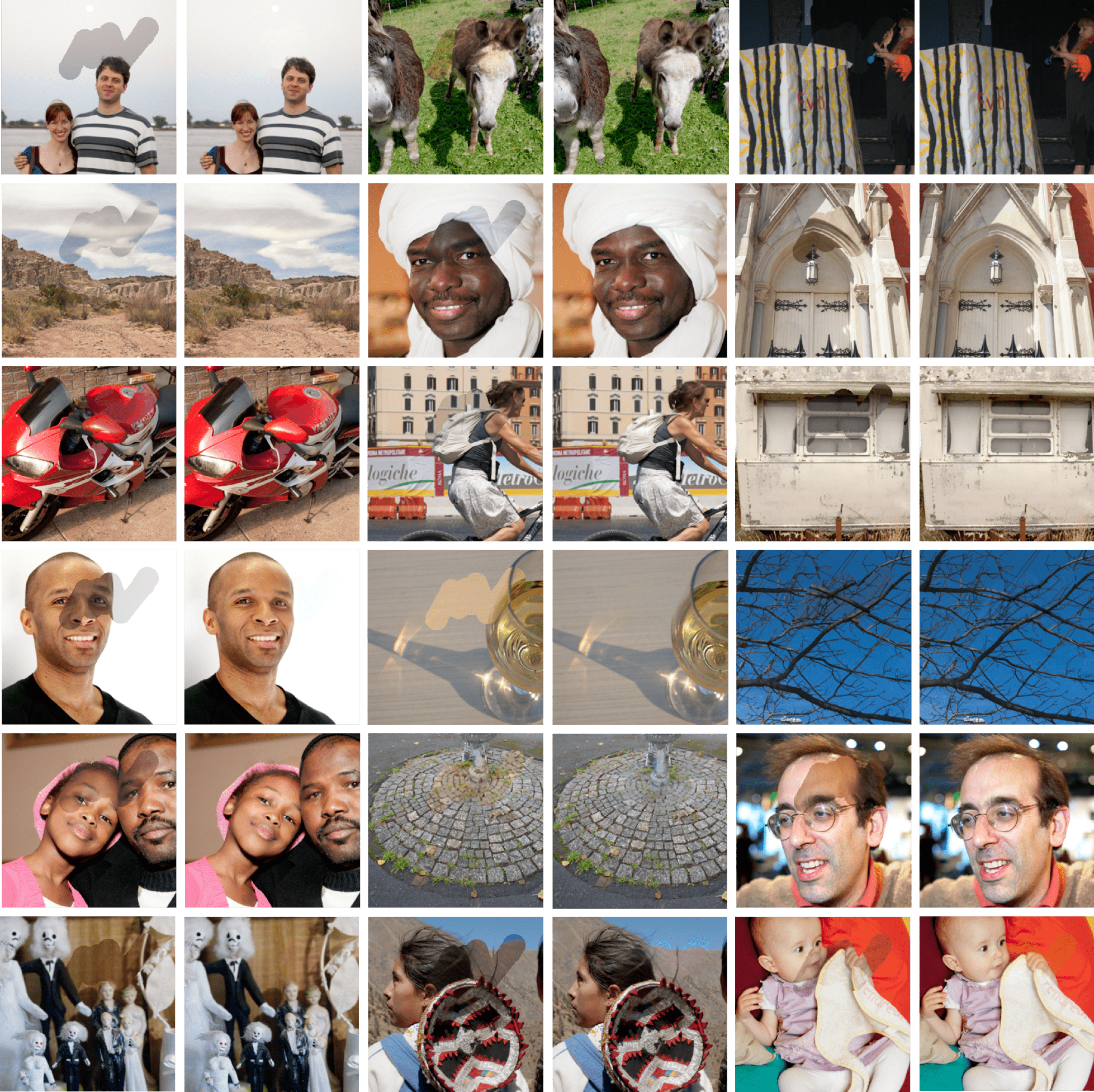}
\caption{Visual results on the synthetic Adobe-5K dataset. For each group of photos, the left one is the composition of $I_t^M$ and $I_s$. We transform the color and warp $I_s$ to make it consistent with $I_t^M$ and composite them as the right image. Our  CST module resolves the color mismatches and spatial misalignment problems. }
  \label{fig:cw}
\end{figure*}

\textbf{Visual Results on Synthetic Adobe-5K}
In Figure \ref{fig:cw}, we show more results on the synthetic Adobe-5K dataset to evaluate the performance of our color transformation. As stated in the paper, we synthesize misaligned and color inconsistent images from Adobe-5K dataset. The spatial transformation is a simple homography-based warping, so the CST module works well to align the images and match the color. More challenging cases can be visualized in user-provided images.

\textbf{Appendix D: More Results on User-provided Images}
Additional higher-resolution results can be found at the following link: \href{https://yzhouas.github.io/projects/TransFill/index.html}{Additional Results}.

\section{Appendix E: Unfolding the Model: Intermediate Results}
\begin{figure*}[t]
\centering
  \includegraphics[width=\linewidth]{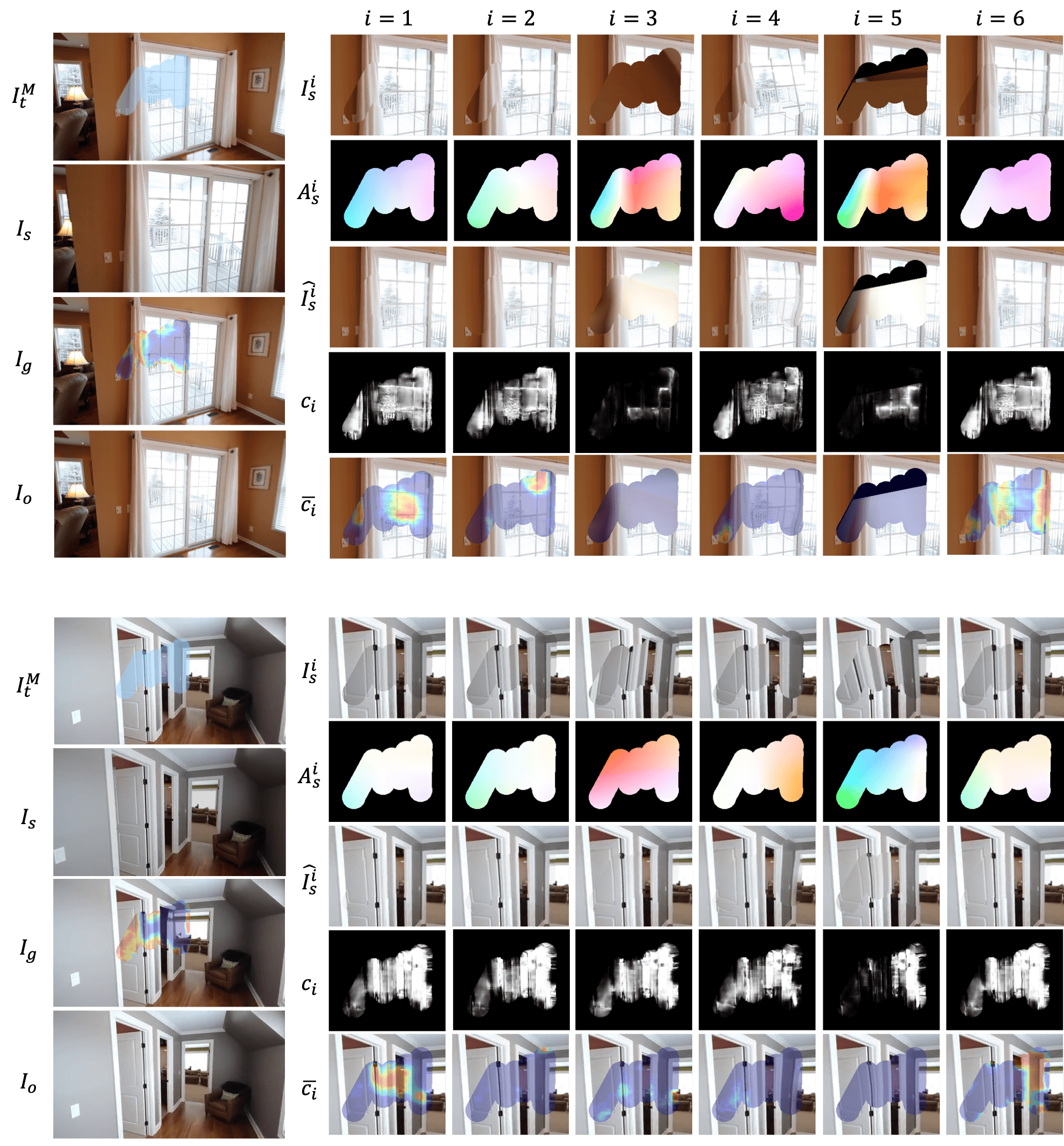}
\caption{Unfolding the whole pipeline to visualize the intermediate results of each module.}
  \label{fig:inter1}
  \vspace{3mm}
\end{figure*}

\begin{figure*}[t]
\centering
  \includegraphics[width=\linewidth]{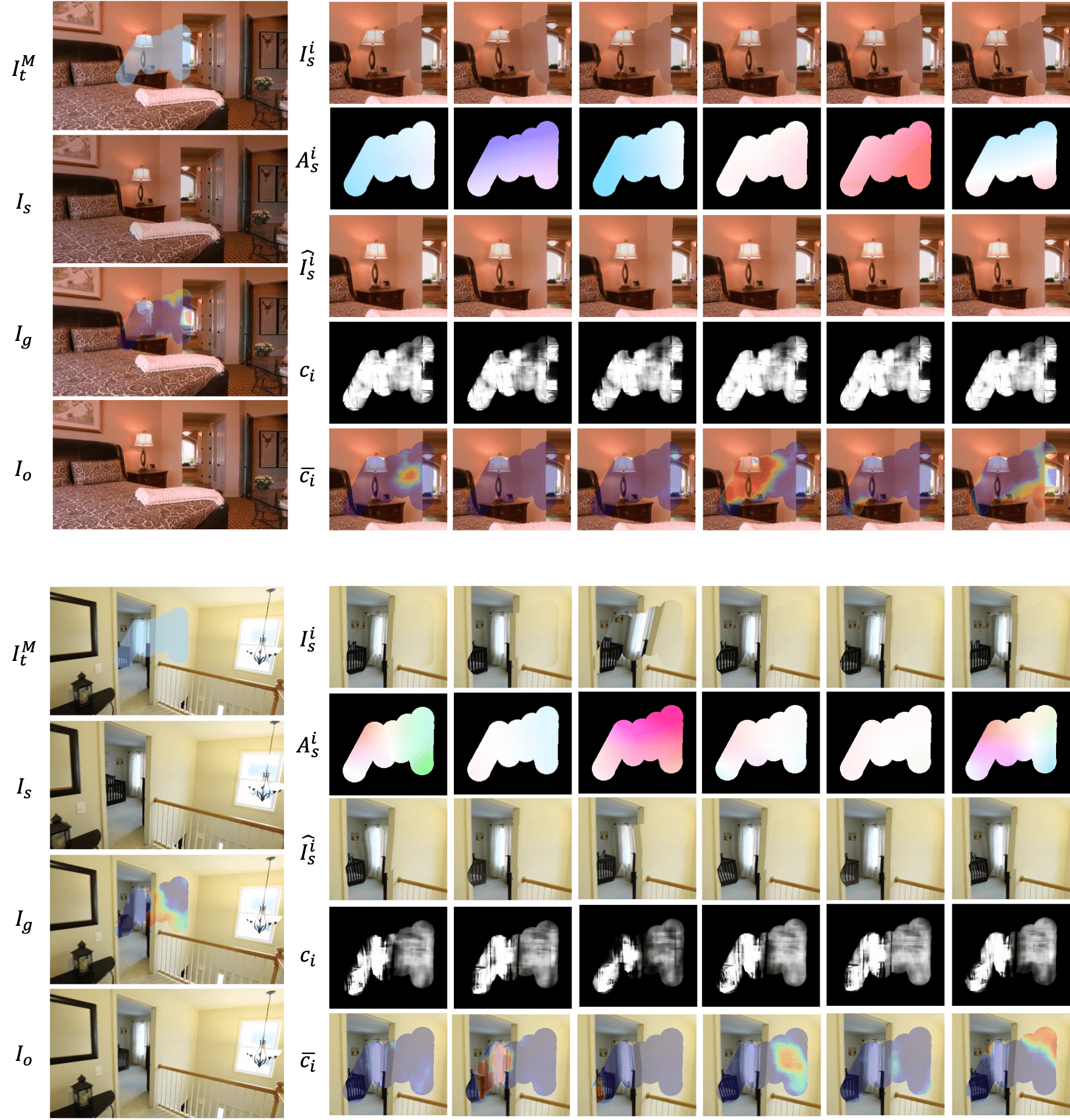}
\caption{Unfolding the whole pipeline to visualize the intermediate results of each module. For some challenging cases when the line alignment is hard, our model can also leverage the outstanding performance of line generation of ProFill to synthesize the door frame. }
  \label{fig:inter2}
  \vspace{3mm}
\end{figure*}

In Figure \ref{fig:inter1} and \ref{fig:inter2}, we unfold the whole pipeline of TransFill to visualize the intermediate results of each proposed module. We demonstrate the process of image completion in a more intuitive way. After proposing different homography-warped images, the CST effectively adjusts the misalignment and color mismatching. Then the proposed TransFill fills in the holes by selectively merging the well-aligned and color-consistent regions from different proposals. Imperfect regions are finally filled with the output from ProFill. 

\clearpage

\clearpage

\clearpage

\section{Appendix F: Network Structures}
The network structures are summarized in Table \ref{table:network1}, \ref{table:network2} and \ref{table:network3}. The structures follow a UNet structure with minor modifications such as some shared structures and some parameter-free components like the tri-linear interpolation layer in the CST. SPF is implemented by a shallower UNet than MPF. 

\begin{table}[h]
\begin{center}
\resizebox{\columnwidth}{!}{
\begin{tabular}{llcl}
\toprule
Input & Output  & \# Out & Type \\ \midrule
$I_t^M, I_s^i, M$ &EncConv1\_1  & 32    & Conv $3\times3$ \\
EncConv1\_1 & EncConv1\_2    & 32    & Conv $3\times3$ \\
EncConv1\_2 & Pool1       & 32    & Maxpool $2\times2$ \\
 Pool1 & EncConv2\_1    & 64    & Conv $3\times3$ \\
EncConv2\_1 & EncConv2\_2    & 64    & Conv $3\times3$ \\
EncConv2\_2 & Pool2       & 64    & Maxpool $2\times2$ \\
Pool2 & EncConv3\_1    & 128   & Conv $3\times3$ \\
EncConv3\_1 & EncConv3\_2    & 128   & Conv $3\times3$ \\
EncConv3\_2 & Pool3       & 128   & Maxpool $2\times2$ \\
Pool3  & EncConv4\_1    & 256   & Conv $3\times3$ \\
EncConv4\_1  & EncConv4\_2    & 256   & Conv $3\times3$ \\
EncConv4\_2 & Pool4       & 256   & Maxpool $2\times2$ \\
Pool4  & EncConv5\_1    & 512   & Conv $3\times3$ \\
EncConv5\_1 & EncConv5\_2    & 512   & Conv $3\times3$ \\
EncConv5\_2  & Pool5       & 512   & Maxpool $2\times2$ \\
EncConv5\_2 & EncConv6\_1    & 512   & Conv $3\times3$ \\
EncConv6\_1 & EncConv6\_2    & 512   & Conv $3\times3$ \\\hline
EncConv6\_2  & ColorCoeff\_1    & 96   & Conv $3\times3$ \\
ColorCoeff\_1  & ColorCoeff\_2    & 96   & Conv $3\times3$ \\\hline
EncConv6\_2  & WarpCoeff\_1    & 2   & Conv $3\times3$ \\
WarpCoeff\_1  & WarpCoeff\_2 +Tanh    & 2   & Conv $3\times3$ \\\hline
$I_s^i$ &GuideConv1\_1  & 16    & Conv $1\times1$\\
GuideConv1\_1&GuideConv1\_2 +Tanh  & 1    & Conv $1\times1$\\
\bottomrule
\end{tabular}}
\end{center}
\caption{Network structure of CST. Prior to each convolution except EncConv1\_1, a PReLU \cite{he2015delving} is applied as a pre-activation.}
\label{table:network1}
\end{table}

\begin{table}[h]
\begin{center}
\resizebox{\columnwidth}{!}{
\begin{tabular}{llcl}
\toprule
Name  & \# Out & Type \\ \midrule
$I_t^M, M, I_s^i$ &EncConv1\_1    & 32    & Conv $3\times3$ \\
EncConv1\_1 & EncConv1\_2    & 32    & Conv $3\times3$ \\
 EncConv1\_2 & Pool1       & 32    & Maxpool $2\times2$ \\
Pool1  & EncConv2\_1    & 64    & Conv $3\times3$ \\
EncConv2\_1  & EncConv2\_2    & 64    & Conv $3\times3$ \\
EncConv2\_2  & Pool2       & 64    & Maxpool $2\times2$ \\
Pool2  & EncConv3\_1    & 128   & Conv $3\times3$ \\
EncConv3\_1 & EncConv3\_2    & 128   & Conv $3\times3$ \\
EncConv3\_2  & Deconv2     & 64   & Deconv $3\times3$ \\
Deconv2 & Concatenate2        & 64    & Deconv2 , EncConv2\_2 \\
Concatenate2  & DecConv2\_1    & 64   & Conv $3\times3$ \\
DecConv2\_1 & DecConv2\_2    & 64   & Conv $3\times3$ \\
DecConv2\_2 & Deconv1     & 32   & Deconv $3\times3$ \\
Deconv1  & Concatenate1        & 32    & Deconv1 , EncConv1\_2 \\
Concatenate1 & DecConv1\_1    & 32   & Conv $3\times3$ \\
DecConv1\_1 & DecConv1\_2    & 32   & Conv $3\times3$ \\
DecConv1\_2  & DecConv1\_3 + Sigmoid    & 1   & Conv $3\times3$ \\\hline
DecConv1\_3  & Concatenate\_feature & 4 & DecConv1\_3, $I_t^M$\\
Concatenate\_feature& FeatureConv1\_1    & 3    & Conv $3\times3$ \\
FeatureConv1\_1 & FeatureConv1\_2    & 3    & Conv $3\times3$ \\ 

\bottomrule
\end{tabular}}
\end{center}
\caption{Network structure of SPF.}
\label{table:network2}
\end{table}

\begin{table}[h]
\begin{center}
\resizebox{\columnwidth}{!}{
\begin{tabular}{llcl}
\toprule
Input & Output  & \# Out & Type \\ \midrule
$I_t^M, M, f_s^i, f_g$   & EncConv1\_1    & 32    & Conv $3\times3$ \\
 EncConv1\_1 & EncConv1\_2    & 32    & Conv $3\times3$ \\
EncConv1\_2  & Pool1       & 32    & Maxpool $2\times2$ \\
Pool1  & EncConv2\_1    & 64    & Conv $3\times3$ \\
EncConv2\_1 & EncConv2\_2    & 64    & Conv $3\times3$ \\
EncConv2\_2 & Pool2       & 64    & Maxpool $2\times2$ \\
Pool2  & EncConv3\_1    & 128   & Conv $3\times3$ \\
EncConv3\_1 & EncConv3\_2    & 128   & Conv $3\times3$ \\
EncConv3\_2 & Pool3       & 128   & Maxpool $2\times2$ \\
Pool3 & EncConv4\_1    & 256   & Conv $3\times3$ \\
EncConv4\_1 & EncConv4\_2    & 256   & Conv $3\times3$ \\
EncConv4\_2 & Pool4       & 256   & Maxpool $2\times2$ \\
Pool4  & EncConv5\_1    & 512   & Conv $3\times3$ \\
EncConv5\_1 & EncConv5\_2    & 512   & Conv $3\times3$ \\
EncConv5\_2 & Deconv4     & 256   & Deconv $3\times3$ \\
Deconv4 & Concatenate4        & 256   & Deconv4 , EncConv4\_2 \\
Concatenate4 & DecConv4\_1    & 256   & Conv $3\times3$ \\
DecConv4\_1  & DecConv4\_2    & 256   & Conv $3\times3$ \\
DecConv4\_2 & Deconv3     & 128   & Deconv $3\times3$ \\
Deconv3 & Concatenate3        & 128   & Deconv3 , EncConv3\_2 \\
Concatenate3 & DecConv3\_1    & 128   & Conv $3\times3$ \\
DecConv3\_1 & DecConv3\_2    & 128   & Conv $3\times3$ \\
DecConv3\_2 & Deconv2     & 64   & Deconv $3\times3$ \\
Deconv2 & Concatenate2        & 64    & Deconv2 , EncConv2\_2 \\
Concatenate2 & DecConv2\_1    & 64   & Conv $3\times3$ \\
DecConv2\_1 & DecConv2\_2    & 64   & Conv $3\times3$ \\
DecConv2\_2 & Deconv1     & 32   & Deconv $3\times3$ \\
Deconv1 & Concatenate1        & 32    & Deconv1 , EncConv1\_2 \\
Concatenate1 & DecConv1\_1    & 32   & Conv $3\times3$ \\
DecConv1\_1 & DecConv1\_2    & 32   & Conv $3\times3$ \\
DecConv1\_2  & DecConv1\_3+Softmax   & $N+2$   & Conv $3\times3$ \\
\bottomrule
\end{tabular}}
\end{center}
\caption{Network structure of MPF.}
\label{table:network3}
\end{table}

\end{document}